\documentclass[english]{IEEEtran}
\usepackage[LGR,T1]{fontenc}
\usepackage[latin9]{inputenc}
\usepackage{color}
\usepackage{array}
\usepackage{textcomp}
\usepackage{mathtools}
\usepackage{multirow}
\usepackage{amsmath}
\usepackage{graphicx}

\makeatletter

\DeclareRobustCommand{\greektext}{%
  \fontencoding{LGR}\selectfont\def\encodingdefault{LGR}}
\DeclareRobustCommand{\textgreek}[1]{\leavevmode{\greektext #1}}
\ProvideTextCommand{\~}{LGR}[1]{\char126#1}

\providecommand{\tabularnewline}{\\}

\usepackage{cite}
\definecolor{red}{rgb}{0,0,0}

\makeatother

\usepackage{babel}
\begin{document}
\markboth{IEEE/ASME Transaction on Mechatronics, Preprint, July 2019}{}

\title{Ceiling Effects for Hybrid Aerial-Surface Locomotion of Small Rotorcraft}

\author{Yi Hsuan Hsiao and Pakpong Chirarattananon, \textit{Member, IEEE}\thanks{Y. H. Hsiao is with the Department of Mechanical Engineering, City University of Hong Kong, Hong Kong SAR, China (email: yhhsiao2-c@my.cityu.edu.hk).}\thanks{P. Chirarattananon is with the Department of Biomedical Engineering, City University of Hong Kong, Hong Kong SAR, China (email: pakpong.c@cityu.edu.hk).}}
\maketitle
\begin{abstract}
As platform size is reduced, the flight of aerial robots becomes increasingly
energetically expensive. Limitations on payload and endurance of these
small robots have prompted researchers to explore the use of bimodal
aerial-surface locomotion as a strategy to prolong operation time
while retaining a high vantage point. In this work, we propose the
use of ``ceiling effects'' as a power conserving strategy for small
rotorcraft to perch on an overhang.\textcolor{red}{{} In the vicinity
of a ceiling, spinning propellers generate markedly higher thrust.
To understand the observed aerodynamic phenomena}, momentum theory
and blade element method are employed to describe the thrust, power,
and rotational rate of spinning propellers in terms of propeller-to-ceiling
distance. The models, which take into account the influence of neighboring
propellers as present in multirotor vehicles, are verified using two
propeller types (\textcolor{red}{23-mm and 50-mm radii}) in various
configurations on a benchtop setup. The results are consistent with
the proposed models. In proximity to the ceiling, power consumption
of propellers with 23-mm radius arranged in a quadrotor configuration
was found to reduce by a factor of three. To this end, we present
a conceptual prototype that demonstrates the use of ceiling effects
for perching maneuvers. Overall, the promising outcomes highlight
possible uses of ceiling effects for efficient bimodal locomotion
in small multirotor vehicles.
\end{abstract}

\section*{Nomenclature}

\begin{tabular}{ll}
$R$ & \textcolor{red}{Propeller's radius}\tabularnewline
$D$ & \textcolor{red}{Propeller-to-ceiling distance}\tabularnewline
$\delta$ & \textcolor{red}{Propeller-to-ceiling ratio ($R/D$)}\tabularnewline
$v,v_{i}$ & \textcolor{red}{Local and induced flow velocities}\tabularnewline
$p,p_{0}$ & \textcolor{red}{Local and atmospheric air pressures}\tabularnewline
$A$ & \textcolor{red}{Area of the propeller disc: $A=\pi R^{2}$}\tabularnewline
$\rho$ & \textcolor{red}{Air density}\tabularnewline
$T$ & \textcolor{red}{Propelling thrust}\tabularnewline
$P_{a},P_{m}$ & \textcolor{red}{Aerodynamic and mechanical powers}\tabularnewline
$\gamma$ & \textcolor{red}{Dimensionless ceiling coefficient}\tabularnewline
$L$ & \textcolor{red}{Propeller-to-propeller distance}\tabularnewline
$\alpha_{0},\alpha_{1}$ & \textcolor{red}{Dimensionless coefficients describing the }\tabularnewline
 & \textcolor{red}{non-axisymmetric flow and wake recirculation}\tabularnewline
$\eta$ & \textcolor{red}{Figure of merit}\tabularnewline
$\Omega$ & \textcolor{red}{Propeller's angular rate}\tabularnewline
$c_{T},c_{\tau}$ & \textcolor{red}{Propeller's thrust and torque coefficients}\tabularnewline
$c_{0,}c_{1,}c_{2}$ & \textcolor{red}{Dimensionless propeller's blade coefficients}\tabularnewline
\end{tabular}

\section{Introduction}

In the past decade, we have witnessed growing developments of Micro
Aerial Vehicles (MAVs). The rapid advancement of these small flying
robots, or drones, is driven by perceivable impacts on a wide range
of applications: transportation and delivery of medical supplies,
environmental monitoring, or enabling ad-hoc network communication
in disaster areas. To date, researchers have demonstrated flight across
various robotic platforms ranging from fixed-wing aircraft with wingspans
of meters \cite{thurrowgood2014biologically,oettershagen2016perpetual},
a swarm of centimeter-scale quadrotors \cite{mulgaonkar2018robust,vasarhelyi2018optimized},
to millimeter-scale flapping-wing robots \cite{chirarattananon2016perching,chen2017biologically}.
Among these, multirotor robots have been widely recognized in both
research and end-user communities owing to the relative ease of use
and expansive functionalities.

However, the high energetic cost of staying airborne poses a major
challenge. The flight time of MAVs is severely constrained by the
onboard power supplies. Compared to fixed-wing counterparts, rotorcraft
encounter an issue of reduction in flight endurance due to the lack
of a large aerodynamically efficient planform. At centimeter scales,
flight at low Reynolds numbers is increasingly difficult thanks to
higher viscous losses \cite{floreano2015science}. The problem aggravates
as the power density of electromagnetic motors decreases and friction
becomes dominant. Flight time of sub-kilogram rotary vehicles swiftly
diminishes to minutes \cite{karydis2017energetics}. This energetic
limitation hinders small flying robots from completing an extended
operation or accomplishing autonomous capabilities that require substantial
payload and power budget for sensors and computation.

Several strategies have been investigated to address the constraints
on payload and flight endurance. In order to expand the navigation
abilities of small robots with limited sensing and computational power,
bio-inspired approaches, such as optic flow-based strategies, have
been explored to enable MAVs to reactively navigate and avoid collisions
\cite{duhamel2013biologically,chirarattananon2018direct}. \textcolor{red}{Control
and planning methods for efficient rotorcraft flight have been proposed
\cite{oktay2015comfortable,morbidi2016minimum}}. Alternatively, to
workaround the elevated costs of aerial transport, researchers have
also demonstrated usages of multimodal locomotion \cite{gnemmi2017conception},
equipping flying robots with abilities to traverse over terrain \cite{daler2015bioinspired},
or perform underwater maneuvers \cite{chen2017biologically,alzu2018loon,siddall2017fast}.
Hybrid aerial and surface locomotion has emerged as another solution
that allows flying robots to perch on surfaces to conserve power while
continuing to carry on functional assignments, such as monitoring,
inspection, or communication \cite{roderick2017touchdown}. To attach
to vertical surfaces or overhangs, various mechanical attachment techniques
and adhesions have been proposed \cite{kalantari2015autonomous,graule2016perching,chirarattananon2016perching,roderick2017touchdown,pope2017multimodal,hang2019perching}.
Common solutions exploit microspines, adhesives, or grasping mechanisms
designed for repeatable attachment and detachment.

This paper investigates the use of ceiling effects for small rotorcraft
to perch on an overhang. \textcolor{red}{With the presence of a ceiling
above a spinning propeller, the structure disrupts the upstream wake.
The airflow in the limited volume between the ceiling and the propeller
lowers the local pressure, effectively attracting the propeller towards
the surface. As a result, the spinning propeller experiences a substantial
increase in aerodynamic force.} The additional thrust potentially
allows a robot to stay aloft at a high vantage point while consuming
less energy. The use of such effects would conceivably enhance operations
of small rotary-wing vehicles in indoor settings or urban environments
with high-rises and elevated structures.

Until now, aerodynamic studies of proximity effects on spinning propellers
are predominantly limited to investigations of ground effects on helicopters.
Based on models using the method of images and a surface singularity,
and experimental validation \cite{betz1937ground,griffiths2005predictions,leishman2006principles,davis2016passive},
the ground effect was found to decrease power consumption by up to
$50\%$ when the propellers are extremely close to the ground \cite{griffiths2005predictions}.
The effects, however, are negligible when the propeller is more than
one diameter above the ground. Small multirotors typically possess
relatively small propellers with an airframe situated below the propellers.
This inevitably enlarges the gap between the ground and the rotors,
rendering the ground effects insignificant \cite{powers2013influence}.
In contrast, we foresee that small MAVs have potential to benefit
from the ceiling effects when operating indoors or under structural
overhangs as the separation between the ceiling and the propellers
can be minimized. Apart from the preliminary findings in \cite{hsiao2018ceiling},
to date, little has been researched on the topic of ceiling effects
as they are irrelevant to flight of traditional helicopters. In the
context of MAVs, a brief study of the ceiling effects was provided
in \cite{powers2013influence}, citing that the resultant force attracts
the vehicle towards the ceiling, increasing the chances of an undesirable
collision. 

This work entails the systematic study of the ceiling effects for
small rotorcraft in terms of force and power. Potential uses of such
effects include the hybrid aerial-surface locomotion for power conservation.
\textcolor{red}{Initially, the impact of a horizontal surface in proximity
to a rotor is analyzed based on classical momentum theory (MT) to
yield the connection between aerodynamic power and thrust.} Despite
requiring some assumptions on the flow conditions, momentum theory
is often employed to describe airflow through wind turbines and propellers
\cite{betz1937ground,leishman2006principles,seddon2011basic,bangura2016aerodynamics,bangura2017thrust,branlard2017wind}.
As first presented in \cite{hsiao2018ceiling}, momentum theory provides
insights into the relationship between the generated thrust and the
aerodynamic power as a function of the propeller-to-ceiling distance.
It turns out that the reduction in aerodynamic power due to the presence
of a ceiling can be quantified using the introduced parameter\textemdash ceiling
coefficient. 

\textcolor{red}{Unlike a single propeller, there might exist flow
interaction between multiple propellers on multirotor robots, we propose
that the values of the ceiling coefficient is also affected by nearby
propellers for the case of multirotor systems owing to asymmetry and
flow recirculation. Next, to gain better insights into how the ceiling
affects the performance of MAVs in flight}, the blade element method
(BEM) is incorporated to relate the thrust and power to the rotational
rate of the blade to obtain the thrust and torque coefficients of
the propellers. The thrust and torque coefficients\textemdash critical
numbers for modeling and flight control applications\textemdash are
no longer constant as in free flight, but dependent on the gap between
the ceiling and the propellers. 

The proposed models are verified by a series of benchtop experiments
on two propeller types in a single and multiple propellers configurations,
with thrust, torque, rotational rate, and power consumptions recorded
for analysis. \textcolor{red}{This allows direct comparison between
the empirically determined ceiling coefficients, and thrust and torque
coefficients against the model predictions.} Lastly, discussion on
the power saving and practical uses of ceiling effects on a small
quadrotor to realize hybrid aerial and surface locomotion is given.

\section{Momentum Theory for Axisymmetric Flow\label{sec:MomentumTheory}}

\begin{figure*}
\begin{centering}
\includegraphics[scale=0.5]{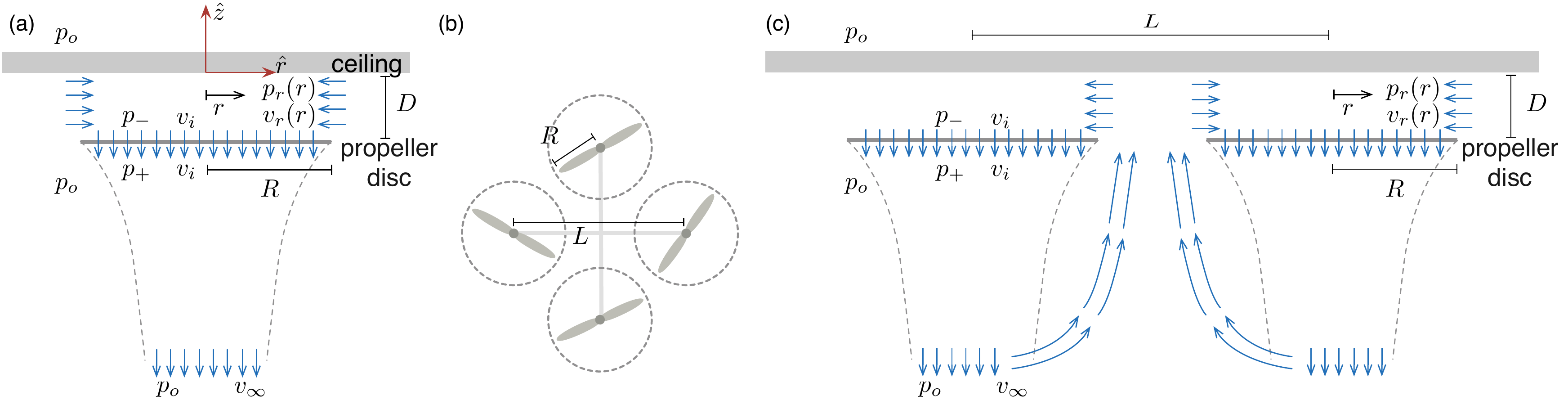}
\par\end{centering}
\caption{A spinning propeller is treated as a thin circular disc situated at
the distance $D$ below the ceiling. The axisymmetric cylindrical
coordinate frame is placed along the axis of the rotor. The diagram
defines the pressure and flow speed at different parts of the system.}
\label{fig:momentum_theory_ceiling_diagram}
\end{figure*}

\textcolor{red}{To quantitatively explain how the presence of a ceiling
alters the power consumption of a spinning propeller}, in this section,
momentum theory is employed to describe the aerodynamic forces and
power associated with a single spinning propeller placed below a flat
surface. To begin, consider the situation described by the diagram
in figure \ref{fig:momentum_theory_ceiling_diagram}(a), an infinitely
thin spinning propeller with radius $R$ is located at the coordinate
$z=-D$ from the horizontal surface at $z=0$. To apply MT, standard
assumptions are used, including that the flow is steady, incompressible
and axisymmetric, the fluid is homogeneous, inviscid, and irrotational,
and the propeller produces thrust by applying the load uniformly through
the actuator disc. Similar to other models that describe rotors or
wind turbines \cite{bangura2016aerodynamics,bangura2017thrust,branlard2017wind},
the flow immediately above and underneath the propeller is assumed
to be one-dimensional. That is, for a stationary propeller, such as
that of a quadrotor in hover, the rotating propeller induces the uniform
vertical airflow, $v_{i}$. This induced velocity is continuous above
and below the propeller disc, consistent with the continuity condition.
However, the actuator disc creates an abrupt change in pressure (from
$p_{-}$ to $p_{+}$ as illustrated in figure \ref{fig:momentum_theory_ceiling_diagram}(a)).
The difference between the downstream and upstream pressures results
in the thrust $T=\left(p_{+}-p_{-}\right)A$. The aerodynamic power
is given by $P_{a}=Tv_{i}$.

The presence of the ceiling affects the airflow above the propeller
as it prevents the upstream air from entering the propeller directly.
When employing the inviscid flow assumption, we drop the no-slip condition
at the wall. It follows that the flow adjacent to the ceiling above
the propeller is radially inward. For a small gap $D$, it is reasonable
to assume that the incoming air at $r=R$ travels horizontally. The
radial component of the flow above the propeller ($v_{r}$) is assumed
to be only dependent on $r$, independent of the distance from the
ceiling. To relate $v_{r}$ to the induced velocity $v_{i}$, apply
the fact that the airflow radially entering an imaginary cylinder
of radius $r$ must vertically exit the cylinder through the propeller
below: $2\pi rD\cdot v_{r}\left(r\right)=\pi r^{2}v_{i}$, or 
\begin{align}
v_{r}\left(r\right) & =\frac{r}{2D}v_{i}\quad\mbox{for }r\in\left[0,R\right].\label{eq:mt_incoming_outgoing_air}
\end{align}
The Bernoulli equation characterizes the relationship between pressure
and velocity of the wake along a streamline \cite{leishman2006principles}.
The upstream air pressures at the ceiling ($p_{r}$) and above the
propeller disc satisfy
\begin{equation}
p_{0}=p_{r}\left(r\right)+\frac{1}{2}\rho v_{r}^{2}\left(r\right)=p_{-}+\frac{1}{2}\rho v_{i}^{2}.\label{eq:mt_bernoulli_upstream}
\end{equation}
Whereas the downstream pressure is related to the terminal flow velocity
such that $p_{+}+\frac{1}{2}\rho v_{i}^{2}=p_{o}+\frac{1}{2}\rho v_{\infty}^{2}.$ 

In regular circumstances, MT states that the thrust force $T$ is
equal to the difference between the vertical momentum of the incoming
and outgoing airflow. In this case, the presence of the ceiling must
be taken into account. Consider an imaginary cylinder of radius $R$
with the top region above the ceiling and the bottom cap infinitely
far away, where the wake has reached terminal velocity, and apply
the conservation of momentum along the vertical direction to objects
and the airflow in this volume. Two external forces include the propelling
thrust ($T$) and the holding force that constrains the ceiling against
the pressure difference above and below the ceiling. The thrust can
be found as 
\begin{align}
T & =\left(p_{+}-p_{-}\right)A=\frac{1}{2}\rho Av_{\infty}^{2}.\label{eq:mt_thrust}
\end{align}
The magnitude of the holding force ($\Delta p\cdot A$) applied to
the ceiling is obtained by integrating $p_{r}\left(r\right)$ over
the surface:
\begin{align}
\Delta p\cdot A & =p_{o}A-\int_{r=0}^{R}p_{r}\left(r\right)2\pi r\mathrm{d}r=\frac{1}{16}\rho Av_{i}^{2}\left(\frac{R}{D}\right)^{2},\label{eq:mt_pressure_diff}
\end{align}
where $p_{r}\left(r\right)$ has been substituted by $v_{r}\left(r\right)$,
and then, $v_{i}$ from equations (\ref{eq:mt_bernoulli_upstream})
and (\ref{eq:mt_incoming_outgoing_air}). In the meantime, it can
be seen that the air enters this imaginary volume horizontally, contributing
to zero vertical momentum. The outgoing flow, with the mass rate $\dot{m}=\rho Av_{i}$,
carries the exit momentum $\dot{m}v_{\infty}=\rho Av_{i}v_{\infty}$.
In total, the conservation of momentum yields
\begin{align}
\frac{1}{2}\rho Av_{\infty}^{2}-\rho Av_{i}v_{\infty}-\frac{1}{16}\rho Av_{i}^{2}\left(\frac{R}{D}\right)^{2} & =0.\label{eq:mt_quadratic_velocities}
\end{align}
\textcolor{red}{The last term in equation (\ref{eq:mt_quadratic_velocities})
distinguishes the considered situation from the no-ceiling case. That
is, the ceiling affects the flow momentum through the pressure difference
above and below the surface.} This equation has one physically feasible
solution:
\begin{equation}
v_{i}=\frac{2}{1+\sqrt{1+\frac{1}{8}\delta^{2}}}\frac{1}{2}v_{\infty}=\frac{1}{2}\gamma^{-1}v_{\infty},\label{eq:mt_vi_vinfty_gamma}
\end{equation}
where a dimensionless ceiling coefficient $\gamma\coloneqq\frac{1}{2}+\frac{1}{2}\sqrt{1+\frac{1}{8}\delta^{2}}$
and the propeller to ceiling ratio $\delta\coloneqq R/D$ are introduced
to capture the effects of the ceiling. When the ceiling is absent
(infinitely far away), $\gamma\rightarrow1$ as found in a regular
case \cite{bangura2016aerodynamics}. Otherwise, $\gamma$ is larger
than unity and monotonically increases as $D$ decreases.

From here, the aerodynamic power is found from $P_{a}=Tv_{i}$ in
terms of $T$ and $\gamma$ using equations (\ref{eq:mt_thrust})
and (\ref{eq:mt_vi_vinfty_gamma}) as
\begin{equation}
P_{a}=\gamma^{-1}T\sqrt{\frac{T}{2\rho A}},\label{eq:mt_power_thrust}
\end{equation}
which implies that, when the ceiling is present ($\gamma>1$), the
propeller requires a factor of $\gamma$ less power to generate the
same thrust. In other words, multirotor vehicles can potentially reduce
power consumption by flying near a ceiling or perching on an overhang.

\section{Efficiency Losses and Multi-Rotor Interaction}

Thus far, the proposed models rely on several assumptions, including
the absence of viscosity and irrotationality, and the symmetry of
the system. It turns out that the simplified model cannot accurately
capture the observed results when multiple rotors operate simultaneously
in a quadrotor configuration as depicted in figure \ref{fig:momentum_theory_ceiling_diagram}(b).
This section examines two primary factors related to the symmetry
of the system and the interaction between adjacent propellers. These
considerations can then be incorporated into the proposed ceiling
coefficient.

\subsection{Irrotational non-axisymmetric flow}

The use of momentum theory in section \ref{sec:MomentumTheory} assumes
quasi-steady flow. The assumption on axisymmetry is somewhat equivalent
to having an infinite number of infinitesimal blades. This, subsequently,
leads to a reasonable time-averaged result. For a more realistic analysis
applicable to a system with finite number of blades, the axisymmetric
assumption is relaxed. That is, the induced velocity $v_{i}$ is allowed
to be dependent on $\theta$ (defined as the angle about the $\hat{z}$
axis in figure \ref{fig:momentum_theory_ceiling_diagram}(a)). Still,
the angular component remains zero. By considering only the first
order variation, it is reasonable to assume
\begin{equation}
v_{i}\left(\theta,\alpha_{0}\right)=v_{i}\left(1+\sqrt{2\left(\alpha_{0}-1\right)}\cos\theta\right),\label{eq:losses_induced_velocity_angular_variation}
\end{equation}
where the term $\cos\theta$, without loss of generality, describes
the first harmonic variation in $\theta$, and $\alpha_{0}$ indicates
the magnitude of the angular variation ($\alpha_{0}\geq1$). When
$\alpha_{0}=1$, $v_{i}\left(\theta,\alpha\right)=v_{i}$ is recovered.
In this form, the average flow velocity is unchanged.

The conservation of mass, as a consequence, requires the radial velocity
of the flow above the propeller to be dependent on $\theta$, changing
equation (\ref{eq:mt_incoming_outgoing_air}) to $v_{r}\left(r,\theta\right)=\frac{r}{2D}v_{i}\left(\theta,\alpha_{0}\right)$.
This alters the average pressure applied to the ceiling. The new holding
force (previously given by equation (\ref{eq:mt_pressure_diff}))
is
\begin{align}
\Delta p\cdot A & =\frac{1}{16}\alpha_{0}\rho Av_{i}^{2}\left(\frac{R}{D}\right)^{2}.\label{eq:losses_ceiling_pressure}
\end{align}
When incorporated into conservation of momentum or equation (\ref{eq:mt_quadratic_velocities}),
the ceiling coefficient becomes $\gamma\left(\delta,\alpha_{0}\right)\coloneqq\frac{1}{2}+\frac{1}{2}\sqrt{1+\frac{\alpha_{0}}{8}\delta^{2}}.$
The angular variation factor, $\alpha_{0}$, can be also be regarded
as an empirical coefficient that accounts for other simplifying assumptions,
such as boundary layer effects, or it can be treated as a factor for
adjusting the effective radius of the propeller blade. The inclusion
of $\alpha_{0}$ does not affect $\gamma$ when the ceiling is absent
or $\delta=0$. 

While in a single propeller case, the time-averaged flow may appear
highly symmetrical. It is likely that at shorter timescales, comparable
to the rotational velocity of the blades, the wake has some angular
variation. This will result in the value of $\alpha_{0}>1$, amplifying
the ceiling coefficient and rendering the propeller more efficient
near a surface. Moreover, it is perceivable that with multiple rotors
operating in proximity, the flow interactions would disrupt the symmetry
of the flow around each propeller, boosting the effective value of
$\alpha_{0}$. These trends are, in fact, observed in the experiments
performed in section \ref{sec:Benchtop-Experiments-and}.

\subsection{Tip loss and recirculation }

With the well-defined streamtube as shown in figure \ref{fig:momentum_theory_ceiling_diagram}(a),
MT and BEM usually neglect tip loss. In reality, the discontinuity
in pressure immediately above and below the propeller disc draws some
downstream air to escape outwards between the blade tips and re-enter
as tip vortices \cite{seddon2011basic}. This essentially reduces
the total induced flow and is known as tip loss. In the case of ground
effect, it is known that the induced velocity is influenced as the
ground prohibits the downward velocity of the wake, resulting in a
lower induced velocity for the same thrust \cite{seddon2011basic}.
The ground also affects the diffusion of tip vortices \cite{light1993tip}.
In \cite{nathan2010rotor}, PIV experiments show that the presence
of the ground causes the recirculation of the wake at a larger scale\textemdash the
effect also known as \textit{brownout}.

For a spinning propeller underneath a horizontal surface, we hypothesize
that part of the wake recirculates in a similar fashion. This phenomenon
is likely more pronounced when multiple propellers are present. As
illustrated in figure \ref{fig:momentum_theory_ceiling_diagram}(c),
when propeller discs are placed at distance $L$ from one another,
their presence obstructs the air from entering the streamtubes above
the propellers. This inevitably leads to recirculation of the wake,
resulting in the reduction of the terminal flow momentum. We postulate
that, depending on the distance between the propeller to the ceiling,
a small portion of the wake (denoted by $\alpha_{1}\delta^{2}$) recirculates.
In this form, the recirculation is more pronounced when the propeller
is closer to the ceiling. As a result, the terminal vertical flow
momentum decreases from $\dot{m}v_{\infty}$ to $\left(1-\alpha_{1}\delta^{2}\right)\dot{m}v_{\infty}$,
whereas the upstream flow and the ceiling pressure are not directly
affected. This modifies the previous conservation of momentum equation
(\ref{eq:mt_quadratic_velocities}). If the earlier angular variation
factor is also considered, the revised ceiling coefficient becomes
\begin{equation}
\gamma\left(\delta,\alpha_{0},\alpha_{1}\right)\coloneqq\frac{1}{2}\left(1-\alpha_{1}\delta^{2}\right)+\frac{1}{2}\sqrt{\left(1-\alpha_{1}\delta^{2}\right)^{2}+\frac{\alpha_{0}}{8}\delta^{2}}.\label{eq:losses_ceiling_coe}
\end{equation}
The recirculation crucially lowers the ceiling coefficient, making
the propeller less energetically efficient. This is because the recirculation
inherently assumes that the downwash loses its energy before re-entering
the upstream wake. In the scenario where multiple rotors are together,
it is anticipated that the neighboring propellers would strengthen
the recirculation, resulting in a larger $\alpha_{1}$ when the distance
between the propellers ($L$) shrinks. Nevertheless, the recirculation
caused by nearby propellers breaks the axisymmetric profile of the
wake. This, in turn, increases $\alpha_{0}$, rendering the propeller
to be more efficient. These two competing phenomena play an important
role in the resultant ceiling coefficient when multiple propellers
operate in a multi-rotor vehicle configuration.

In this form, the ceiling coefficient still asymptotically approaches
unity as the ceiling is infinitely far away, independent of $\alpha_{1}$.
In other words, the parameter $\alpha_{1}$ captures the interaction
between the propellers that are caused by the presence of a ceiling.
However, it does not describe the interactions that may already exist
without the ceiling. 

\subsection{Figure of merit and power\label{subsec:Figure-of-merit}}

At the end of section \ref{sec:MomentumTheory}, the relationship
between the aerodynamic power ($P_{a}$) and thrust has been presented.
This represents the power delivered by the spinning propeller to the
air. The aerodynamic power, however, is less than the mechanical power
delivered by the motor (measurable as the product of torque and angular
velocity, $P_{m}=\tau\Omega$) owing to losses from wake rotation,
non-uniform flow, and tip vortices not captured by momentum theory
\cite{leishman2006principles,bangura2016aerodynamics}. Figure of
merit ($\eta$) accounts for the difference, representing the aerodynamic
efficiency of the rotor: 
\begin{equation}
P_{a}=\eta P_{m}.\label{eq:figure_of_merit}
\end{equation}
This figure of merit is typically lower for smaller rotors as they
are inherently less efficient. For simplicity, it is usually assumed
constant for a particular propeller, regardless of the rotational
rate.

For a motor-propeller system, the input power into the system ($P_{i}=IV$)
is also different from the mechanical power due to heat dissipation
and frictional losses. For a brushed motor, we consider the first-order
motor model in steady state: $V=IR_{i}+V_{e}$, where $R_{i}$ is
the effective motor's internal resistance, and $V_{e}=k\Omega$ is
the back EMF, linearly proportional to $\Omega$. The mechanical power
of the motor is often assumed identical to the electrical power subtracted
by the resistive loss, $P_{m}=IV_{e}=Ik\Omega$. This renders $P_{m}$
to always be lower than $P_{i}$. For a brushless motor driven by
a three phase signal generated by an Electronic Speed Controller (ESC),
the input voltage is approximately constant while the ESC regulates
the current to vary the output power. The mechanical power is equal
to the input power subtracted by losses in the internal resistance
and the ESC. In such cases, it is more sophisticated to determine
the mechanical power from the input voltage and current, however,
it can be calculated from the torque and angular velocity measurements.

\section{Thrust and Torque Coefficients for Flight\label{sec:Thrust-and-Torque}}

\subsection{Blade element momentum theory}

Using MT, the relationship between the generated thrust and aerodynamic
power is given by equation (\ref{eq:mt_power_thrust}). The presence
of a horizontal surface or ceiling above the propeller modifies this
relationship through the ceiling coefficient. \textcolor{red}{MT alone,
however, does not relate the aerodynamic power and thrust to the propeller's
speed or the torque it experiences.} In this section, BEM is used
to consider the geometry of the propeller to estimate the thrust,
torque and power when the propeller spins at the angular rate $\Omega$.
Incorporating this with the ceiling coefficient, the thrust and torque
coefficients of a spinning propeller are evaluated in terms of the
ceiling coefficient. \textcolor{red}{These parameters are important
for flight in the vicinity of a ceiling as the produced force and
torque are affected by the ceiling even when the angular rate is maintained.}

\textcolor{red}{Here, readers are referred to \cite{bangura2016aerodynamics}
and the supplemental materials. In \cite{bangura2016aerodynamics},
blade element momentum theory (BEMT) was used to derive an equation
describing the relationship between thrust, induced velocity and the
angular rate of a propeller. As previously shown in Section \ref{sec:MomentumTheory},
with the presence of a nearby ceiling, the upstream wake acquires
some radial velocity. This is distinct from a regular flight condition,
where a rotor only experiences vertical and horizontal flow. As a
consequence, the previous result from \cite{bangura2016aerodynamics}
must be modified to take into account the contribution from radial
flow across the propeller blades. As derived in the supplemental materials,
}the equation for thrust from BEM becomes
\begin{equation}
T=\frac{1}{2}\rho AR^{2}(c_{0}-c_{1}\frac{v_{i}}{\Omega R}+c_{2}\frac{v_{i}}{\Omega R}\delta)\Omega^{2},\label{eq:bem_ceiling_thrust}
\end{equation}
where the coefficients $c_{0}$, $c_{1}$, and $c_{2}$ are related
to the blade profile. The last term is dependent on the propeller-to-ceiling
ratio, which asymptotically vanishes when the ceiling is infinitely
far away. In this form, the thrust coefficient\footnote{The thrust and torque coefficients ($c_{T}$ and $c_{\tau}$) are
defined as commonly used in analysis of aerial vehicles \cite{bangura2016aerodynamics},
slightly different from the dimensionless definitions used in aerodynamics
research. } $c_{T}\coloneqq T/\Omega^{2}$ cannot be immediately derived from
equation (\ref{eq:bem_ceiling_thrust}) owing to the presence of $v_{i}$.
However, equation (\ref{eq:mt_thrust}) states that $T=\frac{1}{2}\rho Av_{\infty}^{2}$.
With the definition of the ceiling coefficient, this can be written
in terms of the induced velocity $v_{i}$ as

\begin{equation}
T=2\rho A\gamma^{2}v_{i}^{2}.\label{eq:bem_thrust_vi}
\end{equation}
Equating $T$ from (\ref{eq:bem_ceiling_thrust}) and (\ref{eq:bem_thrust_vi})
let us solve for $\Omega R/v_{i}$. When substituted back to equation
(\ref{eq:bem_ceiling_thrust}), the thrust coefficient is obtained,
$c_{T}\coloneqq T/\Omega^{2}$,

\begin{equation}
c_{T}=2\rho A\left(\frac{2c_{0}R\gamma}{\left(c_{1}-c_{2}\delta\right)+\sqrt{\left(c_{1}-c_{2}\delta\right)^{2}+16c_{0}\gamma^{2}}}\right)^{2}.\label{eq:bem_thrust_coef}
\end{equation}
The thrust coefficient depends on the ceiling coefficient and $\delta$.
Without the ceiling, this coefficient reduces to $\left.c_{T}\right|_{\gamma=1}=8\rho A\left[c_{0}R/\left(c_{1}+\sqrt{c_{1}+16c_{0}}\right)\right]^{2}$.

To evaluate the torque coefficient $c_{\tau}\coloneqq\tau/\Omega^{2}$,
we use the fact that $P_{m}=\tau\Omega=c_{\tau}\Omega^{3}$ , $P_{a}=Tv_{i}$,
and $P_{a}=\eta P_{m}$. It follows that
\begin{equation}
c_{\tau}=\frac{1}{\eta\sqrt{2\rho A}}c_{T}^{3/2}.\label{eq:bem_torque_coef}
\end{equation}
Far away from the ceiling $\left.c_{\tau}\right|_{\gamma=1}=\left(16\rho A/\eta\right)\left[c_{0}R/\left(c_{1}+\sqrt{c_{1}+16c_{0}}\right)\right]^{3}$.
As anticipated, without the ceiling and, hence, radial flow, the parameter
$c_{2}$ disappears from the expression of thrust and torque coefficients.

\section{Benchtop Experiments\label{sec:Benchtop-Experiments-and}}

In this section, experiments are conducted to verify the aerodynamic
models for two propeller sizes. The experimental procedure is designed
such that ceiling and propeller coefficients can be empirically evaluated
under various conditions. The results enable the verification of (i)
the relationship between power and thrust at various propeller-to-ceiling
distances; (ii) the impact of multirotor interaction; and (iii) the
resultant thrust and torque coefficients. 

\subsection{Experimental setup}

\begin{figure}
\begin{centering}
\includegraphics[width=7cm]{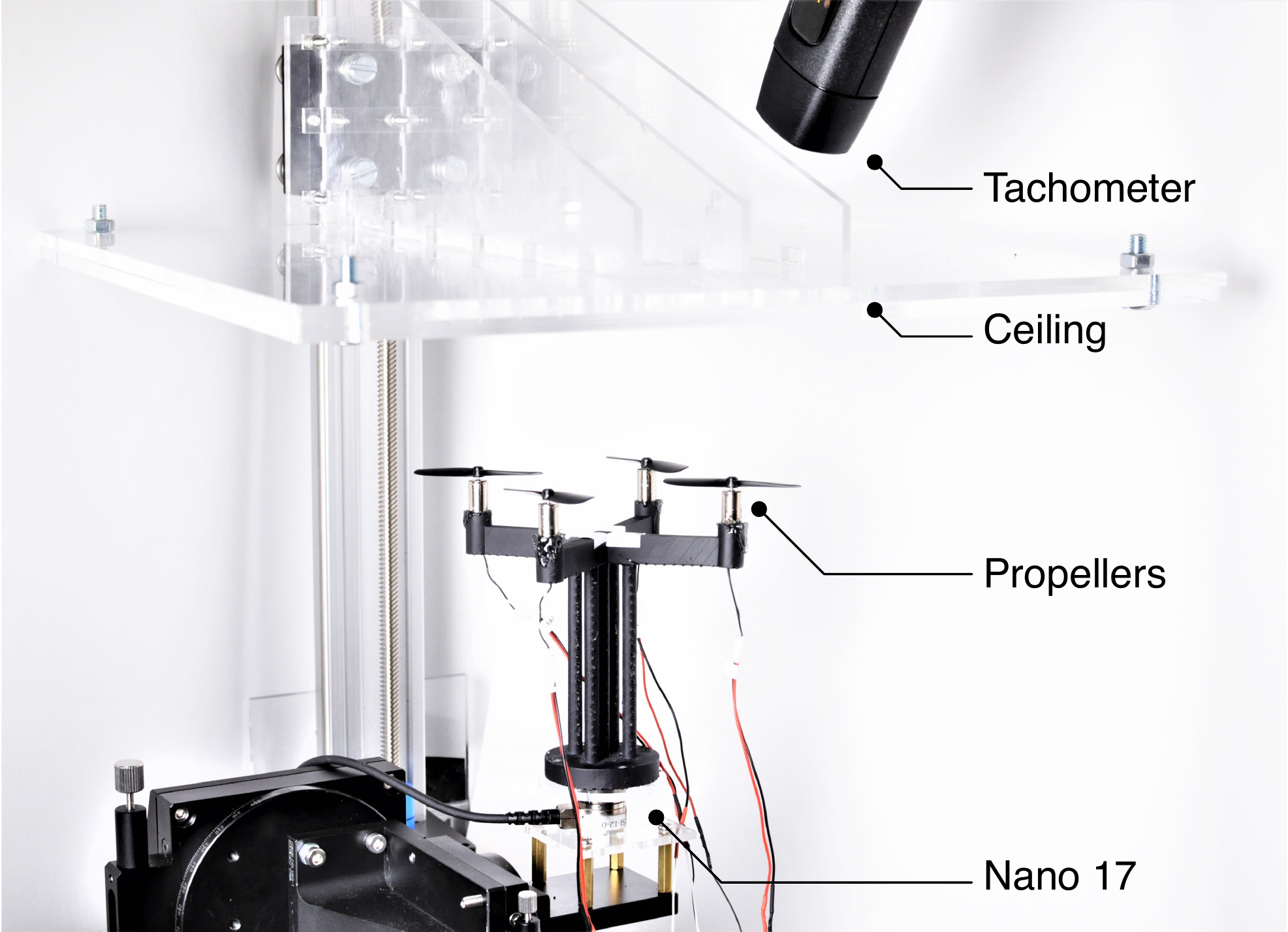}
\par\end{centering}
\caption{A photo depicting the experimental setup with four 23-mm propellers
mounted on a force/torque sensor. The ceiling, which translates vertically,
is controlled by a motorized stage. The tachometer placed above the
ceiling measures the rotational rate belonging to one of the propellers.}
\label{fig:panel_exp_setup}
\end{figure}
Two motor-propeller combinations were employed for the experiments.
Each motor and propeller combination was mounted on a multi-axis force/force
sensor (nano17, ATI) via a 3D printed structure (Black Resin, Formlabs
Form 2) as illustrated in figure \ref{fig:panel_exp_setup}. Two 5
mm-thick transparent acrylic plates (with the total thickness of 10
mm) were mounted on a linear motorized positioning stage as a ceiling.
The stage was driven by a microstepping driver (M542, Leadshine) for
adjusting the distance between the propeller and the ceiling, yielding
a step size of $20$ $\mu$m.

For generation of driving signals and data acquisition, a computer
running the Simulink Real-Time (Mathworks) system with a DAQ (PCI-6229,
National Instruments) was used for generating command signals for
driving the motor and the ceiling, recording force/torque measurements,
and collecting voltage and current data. The Advent Optical A2108
tachometer with an analog output was installed above the transparent
acrylic plates to provide the RPM of the propeller with the accuracy
of $\approx0.5\%$. The RPM measurements are synced with other measurements
via the DAQ at the rate of 1 kHz or higher. Each measurement point
represents the data averaged over two seconds in steady states.

\subsection{Experiments}

\subsubsection{Propeller with a 23-mm radius}

For the first motor-propeller combination, 7 \texttimes{} 16-mm coreless
DC motors and propellers with a 23-mm radius (R = 23 mm) commercially
available as parts for Crazyflie 2.0 were chosen for the experiments.
DC signals between 2.5-4.0 V generated by the DAQ were used as an
input reference for high-current amplifiers (OPA548T, Texas Instruments)
in the voltage follower configuration for directly driving the DC
motors. Current sensors (INA169, Texas Instruments) were incorporated
to measure the current in the range of 0 \textminus{} 5 A with the
errors of 2\%. The voltage across each motor was also monitored through
the DAQ. In the experiments, each motor consumed the maximum of $\approx1.1$
A.

For 23-mm propellers, we first tested a single propeller with the
propeller axis aligned with the $\hat{z}$-axis of the force/torque
sensor. This enabled simultaneous measurements of force and torque.
In steady states, the force represents the axial thrust generated
by the propeller and the axial torque is the aerodynamic drag.

To study the multi-rotor interaction, four propellers were mounted
in a symmetric ``\verb:+:'' configuration using a 3D printed frame.
We experimented with different distances between opposite propellers
($L=$ 78, 85, 92, and 106 mm). In this setting, both clockwise and
counter-clockwise rotating propellers were used to imitate a real
quadrotor. As a consequence, only the force (thrust) measurements
are available.

For each configuration, the propeller-to-ceiling distance was varied
from 1 to 100 mm, resulting in 68 different distances. At each distance,
16 driving voltages were commanded. This resulted in approximately
1,000 measurements for each configuration. In the presentation of
the results, all measurements corresponding to multirotor configurations
are normalized to represent the values per one propeller.

\subsubsection{Propeller with a 50-mm radius}

To show that the proposed model applies generally to not only one
propeller size, we employed a 50-mm radius (R = 50 mm) carbon fiber
propeller paired with a brushless DC motor (MultiStar Viking 2206-2600kv)
for a single propeller experiment. Here, a PWM signal directly generated
by the DAQ was used as command signals for an ESC (Plush 25A, Turnigy)
to produce three-phase signals required by the brushless motor. A
12-V power supply unit (DS550-3, Astec) was used to provide the power
to the motor via the ESC. The supplied voltage across the ESC was
monitored through the DAQ whereas the consumed current was also measured
using a current sensor (GHS 10-SME, LEM USA). In the experiments,
the averaged current varied from $1.5$ to $10.2$ A, while the voltage
remained approximately constant at $\approx12.2-12.3$ V.

With this propeller, experiments with a single propeller were performed
at 68 propeller-to-ceiling distances from 1 to 105 mm. At each distance,
16 values of PWM signals were used. The measurements of RPM, voltage,
current, force, and torque were recorded. In total, over $1,000$
data points were taken.

\section{Experimental Results and Analysis\label{sec:Power,-thrust,-and-1}}

\subsection{Measurement results}

\begin{figure}
\begin{centering}
\includegraphics[scale=0.58]{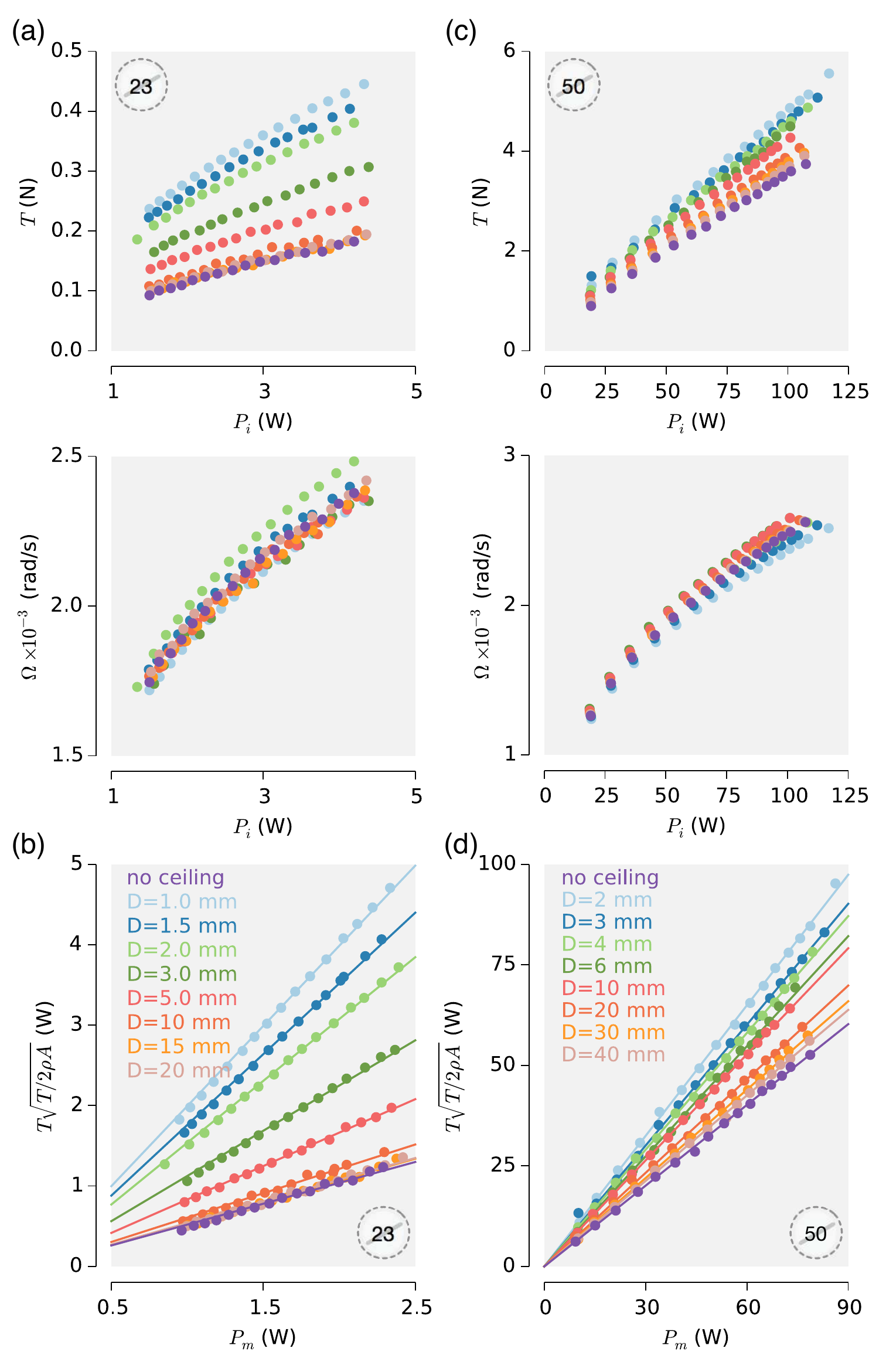}
\par\end{centering}
\caption{Measurements of thrust and rotational rates against input power and
mechanical power for (a) $23$-mm propeller, and (b) 50-mm propeller,
both in a single propeller setting.}
\label{fig:panel_raw_measurements}
\end{figure}
Figure \ref{fig:panel_raw_measurements}(a) shows the measurements
of thrust and angular velocity of a single $23$-mm propeller versus
the input power into the system ($P_{i}$) at some representative
distances from the ceiling (other measurements are omitted for clarity).
Overall, as the driving voltage increases, the current and therefore
power, rise as expected. This results in higher thrust forces and
angular velocities. Without the ceiling, the maximum thrust value
is 0.18 N. This value increases dramatically with the ceiling, reaching
$0.45$ N ($2.5$ times of $0.18$ N) when the ceiling is $1.0$ mm
from the propeller ($D=1.0$ mm, $\delta=23.0$) while the input power
remains approximately unchanged. In terms of the rotational rate,
relatively little changes are observed when the ceiling is introduced.
The angular velocity is primarily dependent on the driving voltage
rather than the distance to the ceiling. Measurements corresponding
to multirotor configurations, which are not shown, are found to feature
similar characteristics.

In case of the $50$-mm propeller, similar trends are seen in figure
\ref{fig:panel_raw_measurements}(c). The effects of the ceiling are
visible, but not as pronounced. For example, the maximum thrust force
increases from 3.7 N without the ceiling to $5.6$ N (1.5 times of
$3.7$ N) when the ceiling is 2 mm away ($D=2$ mm, $\delta=25.0$).
Similarly, the presence of the ceiling only slightly affects the angular
velocity.

\subsection{Calculation of mechanical power}

To determine the ceiling coefficients, it is required to evaluate
the mechanical power outputted by the motors. For a single 23-mm propeller
with a coreless motor, the mechanical power can be deduced as $P_{m}=\tau\Omega$
from the measurements of torque and angular velocity. To determine
the mechanical power for multi-rotor configurations, where the propeller
torques cancel out, the first-order motor model $V=IR_{i}+k\Omega$
or $P_{i}=IV=I^{2}R_{i}+P_{m}$ as outlined in section \ref{subsec:Figure-of-merit}
is considered. With the knowledge of $P_{m}=\tau\Omega$ from a single
propeller case (where the measurements of $\tau$ is available) and
measurements of $I$ and $V$, we solve for $R_{i}$ from the equation
$P_{i}=I^{2}R_{i}+P_{m}$ using the least-squares method. Assuming
this $R_{i}$ is identical for all motors, the least-squares method
is applied to determine $k$ from $V=IR_{i}+k\Omega$ based on the
measurements of $V,$ I, and $\Omega$. The resultant $R_{i}$ and
$k$ are found to be $1.58$ $\Omega$ and $1.1$ mV$\cdot$s$\cdot$rad$^{-1}$.
The mechanical power is then given as $P_{m}=Ik\Omega$.
\begin{figure}
\begin{centering}
\includegraphics[scale=0.58]{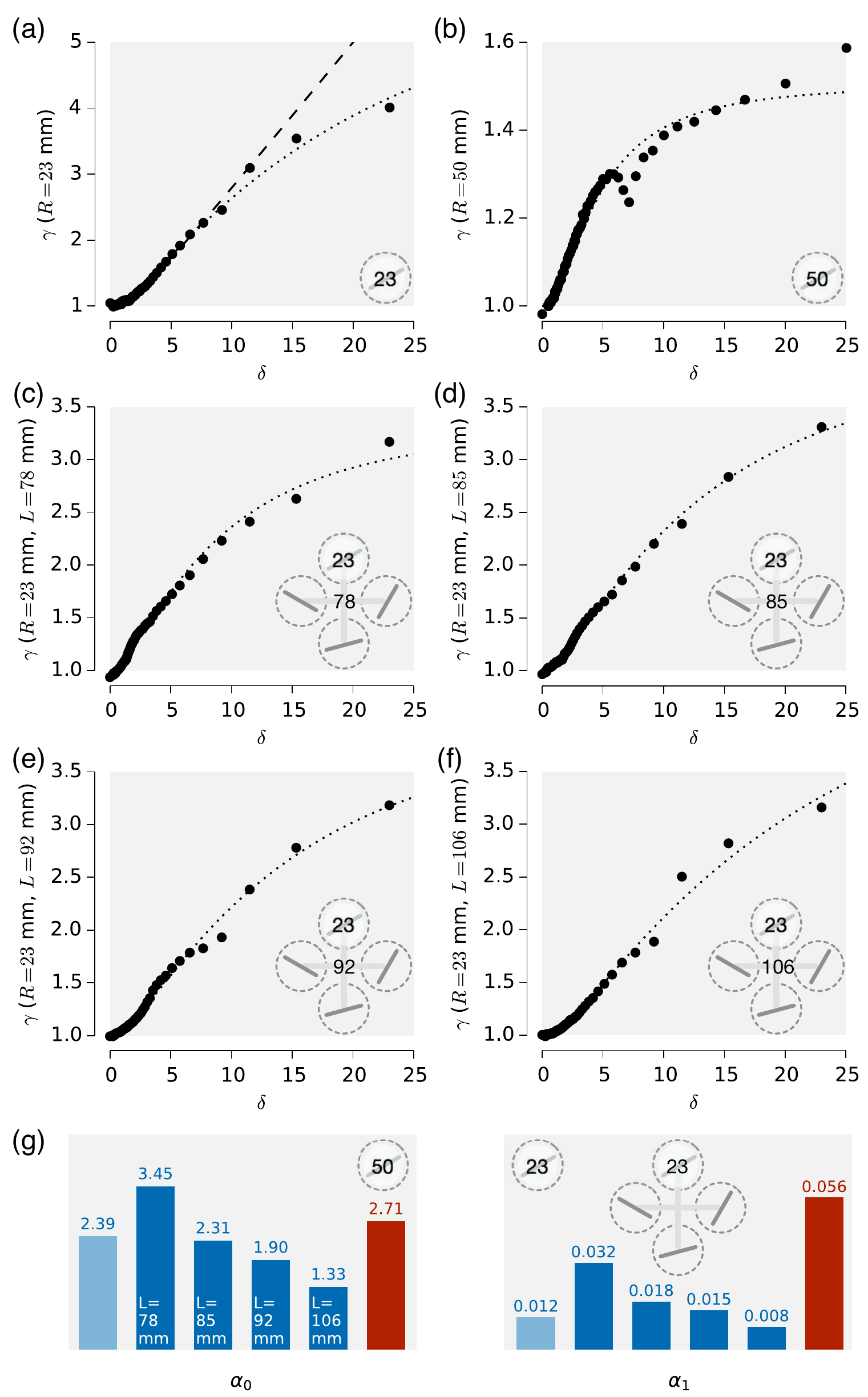}
\par\end{centering}
\caption{Empirical ceiling coefficients (points) compared to the proposed model
(dashed line for the reduced model and dotted lines for the full model).
(a) a single 23-mm propeller. (b) a single 50-mm propeller. (c)-(f)
four 23-mm propellers arranged in a quadrotor configuration with $L=78$,
$\text{85}$, $92$, and $105$ mm. (g) Fitted model coefficients
for six tested propeller configurations. }
\label{fig:panel_ceiling_coefs}
\end{figure}

In case of the $50$-mm propeller, only experiments with a single
propeller were performed. The mechanical power is, therefore, readily
available as $P_{m}=\tau\Omega$. The measurements of $V$ and $I$
provide the input power only for reference and comparison.

\subsection{Ceiling coefficients}

\subsubsection{Ceiling coefficients from the experiments}

From the calculated mechanical power, we plot $P_{m}$ against $T\sqrt{T/2\rho A}$
as suggested by equation (\ref{eq:mt_power_thrust}) and (\ref{eq:figure_of_merit})
using $\rho=1.2$ kg.m$^{-3}$. For each particular distance from
the ceiling, the data points from different commanded signals amount
to the best fit line, of which the slope corresponds to the inverse
of the ceiling coefficient times the figure of merit ($P_{m}=\frac{1}{\eta\gamma}T\sqrt{T/2\rho A}$).
This allows $\eta$ and $\gamma$ to be empirically deduced from the
power and thrust measurements. Examples of the plots are given in
figure \ref{fig:panel_raw_measurements}(b),(d) for single propeller
tests for both $23$-mm and 50-mm propellers. It can be seen that
the relationship between $P_{m}$ and $T\sqrt{T/2\rho A}$ is linear
as anticipated. The corresponding figures of merit for the $23$-mm
and $50$-mm propellers are $\eta=0.50$ and $0.68$. The values imply
that the smaller propeller is aerodynamically less efficient as expected
\cite{bangura2016aerodynamics,bangura2017thrust}. Moreover, the ceiling
coefficients deduced from the gradients increase as the distance to
the ceiling reduces. The trend is visibly more prominent for the 23-mm
propeller. The plots for other propeller configurations are omitted
for brevity.

The ceiling coefficients found from the experimental data (the gradients
of fitted lines in figure \ref{fig:panel_raw_measurements}(b),(d))
for all propeller configurations are plotted against $\delta$ as
points in figure \ref{fig:panel_ceiling_coefs}(a)-(f). It can be
seen that the presence of the ceiling boosts the values of $\gamma$
significantly. For a single $23$-mm propeller, the ceiling coefficients
increase from unity to $\gamma\approx4$ when $\delta=23$ or $D=1.0$
mm. The implies that, at $1.0$ mm from the ceiling, the thrust is
amplified by a factor of $4^{2/3}$ or $2.5$ times for the same power
consumption, consistent with the observations in figure \ref{fig:panel_raw_measurements}.
The ceiling effects, however, appear less prominent for propellers
in quadrotor configurations. The coefficients for the multirotor cases
maximize around $\approx3-3.5$ and are qualitatively similar for
all $L$'s. The effects of the ceiling are also evident for a 50-mm
propeller, nevertheless, the increase in thrust is notably smaller
than that of the 23-mm propeller. At $\delta=25$ or $D=2$ mm, $\gamma$
is approximately $1.6$, suggesting a $37\%$ improvement in thrust
given the same power consumption. Furthermore, we observe an anomalous
dip in the values of $\gamma$ near $\delta\approx7.1$. In fact,
similar features are also perceived with 23-mm propellers near $\delta\approx9.2$,
to a smaller extent. \textcolor{red}{We believe this is caused by
unmodeled power losses as discussed further in section \ref{subsec:Fitted-models-for}
and the supplemental materials}.

\subsubsection{Proposed models of the ceiling coefficients}

Next, the empirically computed ceiling coefficients are taken to evaluate
the best fitted coefficients ($\alpha_{0}$ and $\alpha_{1}$) of
the proposed model as described in equation (\ref{eq:losses_ceiling_coe}).
First, $\alpha_{1}$ is first assumed to be zero. In other words,
the wake recirculation is neglected. This reduced model is similar
to our preliminary findings presented in \cite{hsiao2018ceiling}.
The reduced model is found to be sufficiently accurate to describe
the observed ceiling effects for a single 23-mm propeller when $\delta<20$.
The prediction of the ceiling coefficients compared to the empirical
results are shown as the dashed line and points in figure \ref{fig:panel_ceiling_coefs}(a)
with $\alpha_{0}=1.60$. Nevertheless, we find that the reduced model
overestimates the ceiling coefficients at higher $\delta$ or when
it is applied to other propeller settings. The results indicate that
the flow recirculation, which captures the partial loss of terminal
flow momentum, must be taken into account.

The dotted lines in figure \ref{fig:panel_ceiling_coefs}(a)-(f) represent
the fitted models with recirculation based on equation (\ref{eq:losses_ceiling_coe})
for both propeller sizes at different configurations. The corresponding
numerical coefficients accounting for the asymmetric flow and recirculation
($\alpha_{0}$ and $\alpha_{1}$) for all configurations are shown
in figure \ref{fig:panel_ceiling_coefs}(g). Apart from the unexpected
dip in $\gamma$ at $\delta\approx7.1$ and $9.2$ mentioned earlier,
the proposed models accurately describe the experimental results in
both single and multi-rotor settings for 23-mm propellers, whereas
in the case of the larger propeller, the model slightly underpredicts
the ceiling coefficient when the propeller is extremely close to the
ceiling ($\delta=24$, $D=2$ mm).

A closer inspection of figure \ref{fig:panel_ceiling_coefs}(g) reveals
that, the asymmetric flow parameters ($\alpha_{0}$) are all above
unity, suggesting some degree of asymmetry in all settings. In particular,
it can be seen that, for quadrotor-like configurations, the value
of $\alpha_{0}$ grows as the distance between the propeller shrinks,
consistent with the assumption of flow interaction. Simultaneously,
we observe an increase in the recirculation factor ($\alpha_{1}$)
as $L$ reduces. The observations are reasonable as the presence of
other rotors in vicinity would introduce recirculation and enhance
the asymmetrical flow pattern as we speculated earlier. The competing
effects of $\alpha_{0}$ and $\alpha_{1}$ make the resultant ceiling
coefficients for all multirotor configurations qualitatively similar
as seen in figure \ref{fig:panel_ceiling_coefs}(c)-(f).

Compared to the small propellers, the fitted coefficients for the
$50$-mm propeller signify markedly higher flow recirculation (see
figure \ref{fig:panel_ceiling_coefs}(g)). Based on this finding,
recirculation is a major factor that demotes the ceiling effects for
the large propeller. It is also likely that, at extremely small gap
sizes ($\delta>20$), the presence of the ceiling starts to disrupt
the recirculation, resulting in larger ceiling coefficients than predicted
by the fitted model, explaining the observation in figure \ref{fig:panel_ceiling_coefs}(b).
\begin{figure}
\begin{centering}
\includegraphics[scale=0.58]{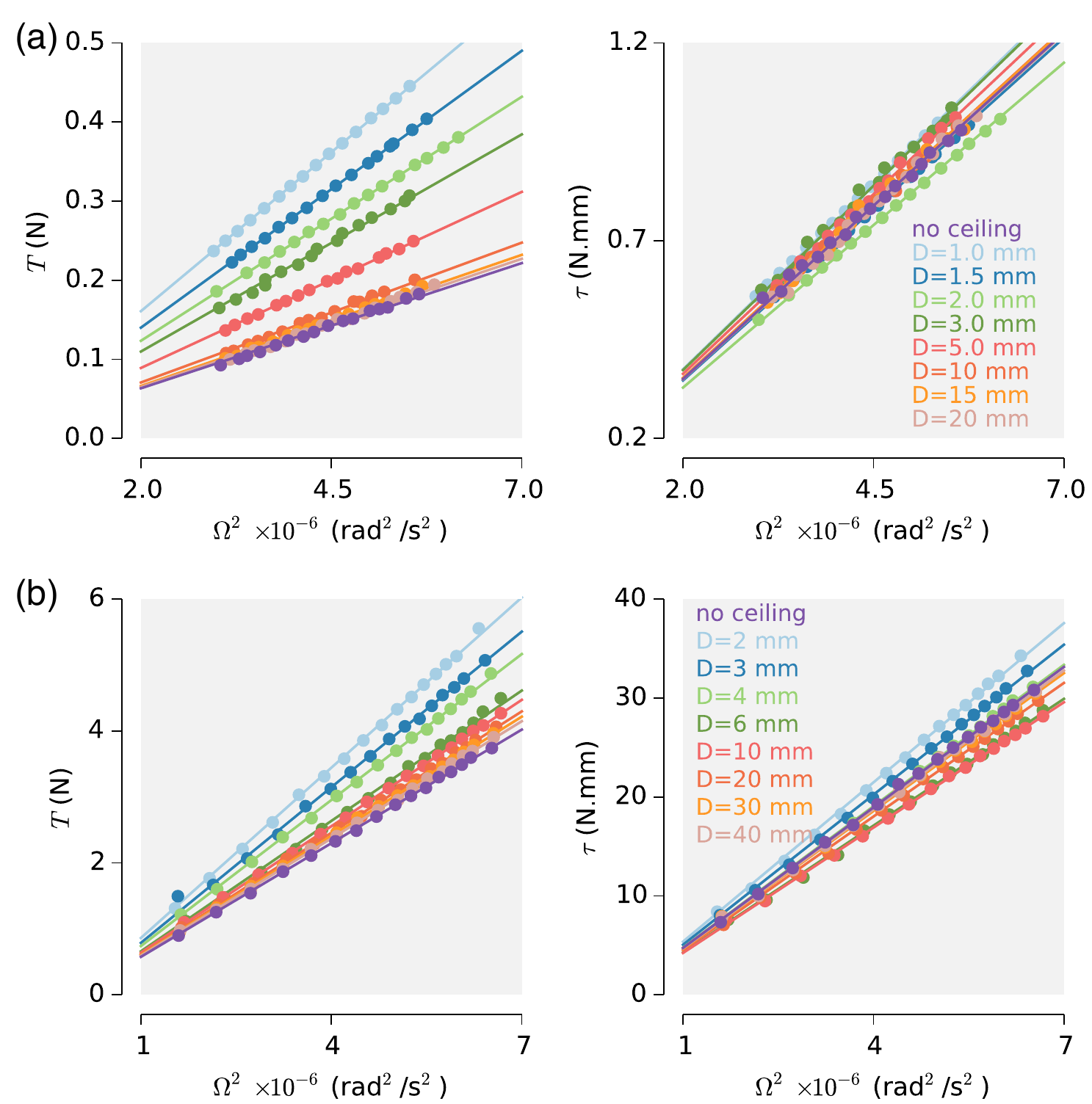}
\par\end{centering}
\caption{Raw measurements of thrust and torque plotted with respect to the
squared rotational rates at different propeller-to-ceiling distances.
(a) Measurements from a single 23-mm propeller. (b) Measurements from
a single 50-mm propeller. }
\label{fig:panel_exp_flight_coeffs}
\end{figure}

\subsection{Thrust and Torque Coefficients for Flight\label{sec:Thurst-and-Torque}}

\subsubsection{Thrust and torque coefficients from the experiments}

The measurements of thrust, torque, and angular velocity enable the
calculation of thrust and torque coefficients of the propellers as
$c_{T}\coloneqq T/\Omega^{2}$ and $c_{\tau}\coloneqq\tau/\Omega^{2}$
. Without a ceiling, these coefficients are constant and only dependent
on the propeller profiles. Without precise knowledge of the blade
profile, $c_{T}$ and $c_{\tau}$ are typically experimentally determined
for flight control purposes. With a ceiling in proximity, the BEMT
analysis suggests that these coefficients also depend on $\delta$
and $\eta$ as given by equations (\ref{eq:bem_thrust_coef}) and
(\ref{eq:bem_torque_coef}).

Focusing on single propeller cases, of which the torque measurements
are available, $T$ and $\tau$ are plotted against $\Omega^{2}$
to compute $c_{T}$ and $c_{\tau}$ corresponding to different ceiling
distances from the gradients. Example data from some representative
distances are illustrated in figure \ref{fig:panel_exp_flight_coeffs}.
The linear relationships between $T$ and $\tau$ with respect to
$\Omega^{2}$ qualitatively verify the validity of equations (\ref{eq:bem_thrust_coef})
and (\ref{eq:bem_torque_coef}), which can be interpreted as, there
exist constant values of $c_{T}$ and $c_{\tau}$ for a fixed propeller-to-ceiling
distance.
\begin{figure}
\begin{centering}
\includegraphics[scale=0.58]{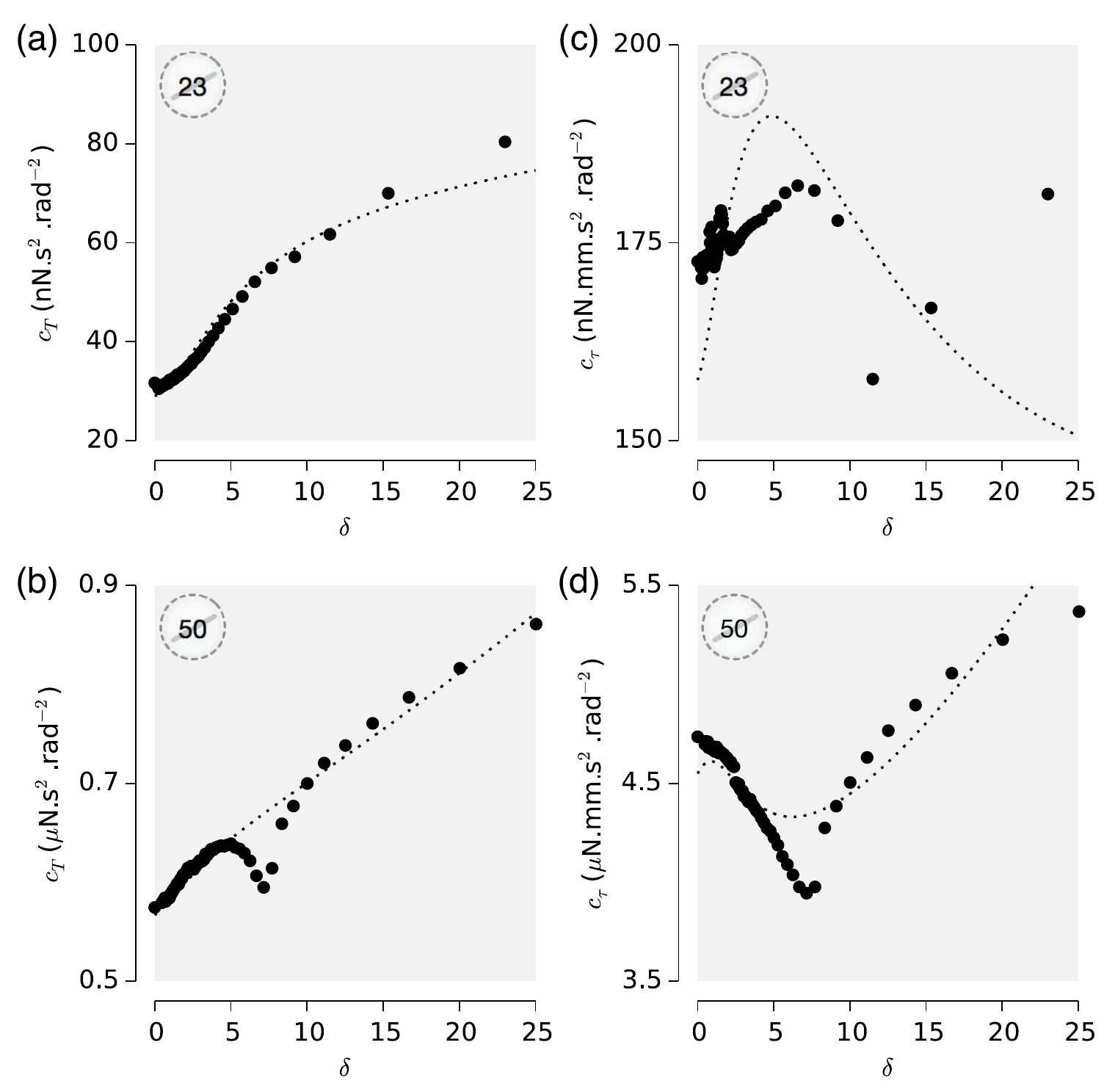}
\par\end{centering}
\centering{}\caption{Flight coefficients as calculated from the benchtop experiments (points)
and from the fitted models (dotted lines). (a) Thrust coefficient
of a 23-mm propeller. (b) Torque coefficient of a 23-mm propeller.
(c) Thrust coefficient of a 50-mm propeller. (d) Torque coefficient
of a 50-mm propeller.}
\label{fig:panel_model_flight_coeffs}
\end{figure}

\subsubsection{Fitted models for thrust and torque coefficients\label{subsec:Fitted-models-for}}

\begin{table}
\begin{centering}
\caption{Propeller coefficients}
\label{tab:flight_coefficients}
\par\end{centering}
\centering{}%
\begin{tabular}{c|ccc|cc}
\hline 
\multirow{2}{*}{Propellers} & $c_{0}$ & $c_{1}$ & $c_{2}$ & $\left.c_{T}\right|_{\delta=0}$  & $\left.c_{\tau}\right|_{\delta=0}$\tabularnewline
 &  &  &  & (Ns$^{2}$rad$^{-2}$) & (Nms$^{2}$rad$^{-2}$)\tabularnewline
\hline 
$23$ mm & 0.154 & 0.846 & 0.022 & 29.0$\times10^{-9}$ & 158$\times10^{-12}$\tabularnewline
$50$ mm & 0.058 & 0.095 & 0.011 & 0.57$\times10^{-6}$ & 4.55$\times10^{-9}$ \tabularnewline
\hline 
\end{tabular}
\end{table}
The analytical expressions of $c_{T}$ and $c_{\tau}$ are functions
of $\delta$ and the dimensionless parameters: $c_{0}$, $c_{1}$,
and $c_{2}$, according to equations (\ref{eq:bem_thrust_coef}) and
(\ref{eq:bem_torque_coef}). Since it is not practical to calculate
these parameters for propellers with sophisticated profiles, the values
of $c_{0}$, $c_{1}$, and $c_{2}$, that best fit our experimental
data from figure \ref{fig:panel_exp_flight_coeffs} are numerically
determined for both $c_{T}$ and $c_{\tau}$. The parameters for both
propellers are listed in table \ref{tab:flight_coefficients}. The
thrust and torque coefficients corresponding to these parameters are
plotted (dotted lines) alongside the empirical results (points) in
figure \ref{fig:panel_model_flight_coeffs}. 

Overall, the fitted models agree with the experimental data for all
$\delta$'s. In the case of the 23-mm propeller, the model for the
thrust coefficient correctly predicts the magnification of more than
$2.5$ times when the ceiling is 1.5 mm (\textgreek{d} \ensuremath{\approx}
15) away from the propeller, with a slight deviation at $D=1.0$ mm
($\delta=23$). The model appears similarly accurate in estimating
the thrust coefficients for the $50$-mm propeller, with the exception
being when $\delta$ is near 9.2, where the irregular drop in the
values of $\gamma$ is observed (refer to figure \ref{fig:panel_ceiling_coefs}(b)). 

\textcolor{red}{While the anomaly cannot be directly explained by
the proposed models, during the experiments, the 50-mm propeller generated
abnormally loud noises when $\delta$ is near 9.2. As suggested by
the analysis of mechanical resonance given in the supplemental materials,
it is highly possible that in this region, the propeller rotated at
frequencies near the standing wave frequency of the experimental setup.
The condition may have affected the wake, resulting in the oscillation
that renders MT inaccurate at predicting the flow dynamics.}

For torque coefficients, at the first glance, the discrepancy between
the predictions of $c_{\tau}$ and the data seems substantial, particularly
for the 23-mm propeller.\textcolor{red}{{} One possible reason is the
calculation of $c_{\tau}$ assumes the figure of merit remains constant.
This may not be entirely accurate as the ceiling might have influenced
the rotational component of the wake.} Despite that, a closer inspection
reveals that the model is sufficiently accurate as it predicts minor
changes in $c_{\tau}$, on the order of 20\% compared to the no ceiling
case. Over the range of $\delta<20$, the differences between the
empirical $c_{\tau}$ and predicted $c_{\tau}$ are less than $\approx10\%$.
For the small propeller, the model correctly predicts the existence
of a peak in $c_{\tau}$ (though the peak location is slightly misaligned
from the experimental results). In total, we see that the proposed
model is able to describe two qualitatively distinct $c_{\tau}$ profiles
obtained from different propeller sizes.

The findings on $c_{T}$ and $c_{\tau}$ here are consistent with
the initial measurements from figure \ref{fig:panel_raw_measurements}.
The fact that the presence of a ceiling only marginally affects $c_{\tau}$
means that it only has minor influences on the power. Consequently,
we observe little changes in the rotational rates given the input
power. In the meantime, the dramatic change in $c_{T}$ is consistent
with the marked changes in resultant thrust at different ceiling distances
for the same input power.

\section{Discussion and Future Work\label{sec:Discussion-and-Future}}

With foreseeable potential as an energy conserving strategy for small
flying robots, this paper studied the effects of ceiling in proximity
to a small spinning propeller. Based on a few simplifying assumptions,
momentum theory and the blade element method were employed to derive
analytical models that describe the thrust, power, and rotational
velocity of a spinning propeller. The formulations were extended to
take into account the presence of nearby propellers to imitate the
propellers in multirotor vehicles. This was achieved by consideration
of wake recirculation and asymmetrical flow pattern. Benchtop experiments
involving propellers with 23-mm and 50-mm radii in single and multi-rotor
settings were performed and the results obtained are consistent with
our model predictions.

As suggested by the model, we found that the ceiling can radically
affect the power consumption and thrust generated by propellers. For
small propellers arranged in a quadrotor-like configuration, we observe
a reduction in mechanical power by a factor of three or more. Whereas
for a 50-mm propeller, the change in power efficiency is lower, the
improvement of approximately $50\%$ is still substantial. While the
current study is still limited to flat, rigid, and horizontal ceilings,
we believe our promising outcomes here offer an opportunity to alleviate
the issue of diminished flight endurance of small rotorcraft.

It is, however, challenging to realize a flight with surface locomotion
in practice. The strategy necessitates the design of a lightweight
mechanism that impedes the robot from directly colliding with the
ceiling, absorbs the kinetic energy from impact to prevent bouncing,
and maintains a suitable propeller-to-ceiling distance. In terms of
flight control, a controller must be devised to deal with the ceiling
approach (which could be different from a regular landing maneuver).
Not only must the controller be able to regulate the thrust appropriately
once the robot is in contact with the ceiling, but it also has to
retain the attitude and spatial stability with the presence of the
normal force from the ceiling. For these reasons, the realization
of surface flight is beyond the scope of this work. Still, an extended
analysis on the power saving potential and a preliminary design of
a lightweight mechanism that would allow the robot to safely approach
the ceiling for the proposed surface locomotion are provided below.
\begin{figure}
\begin{centering}
\includegraphics[scale=0.58]{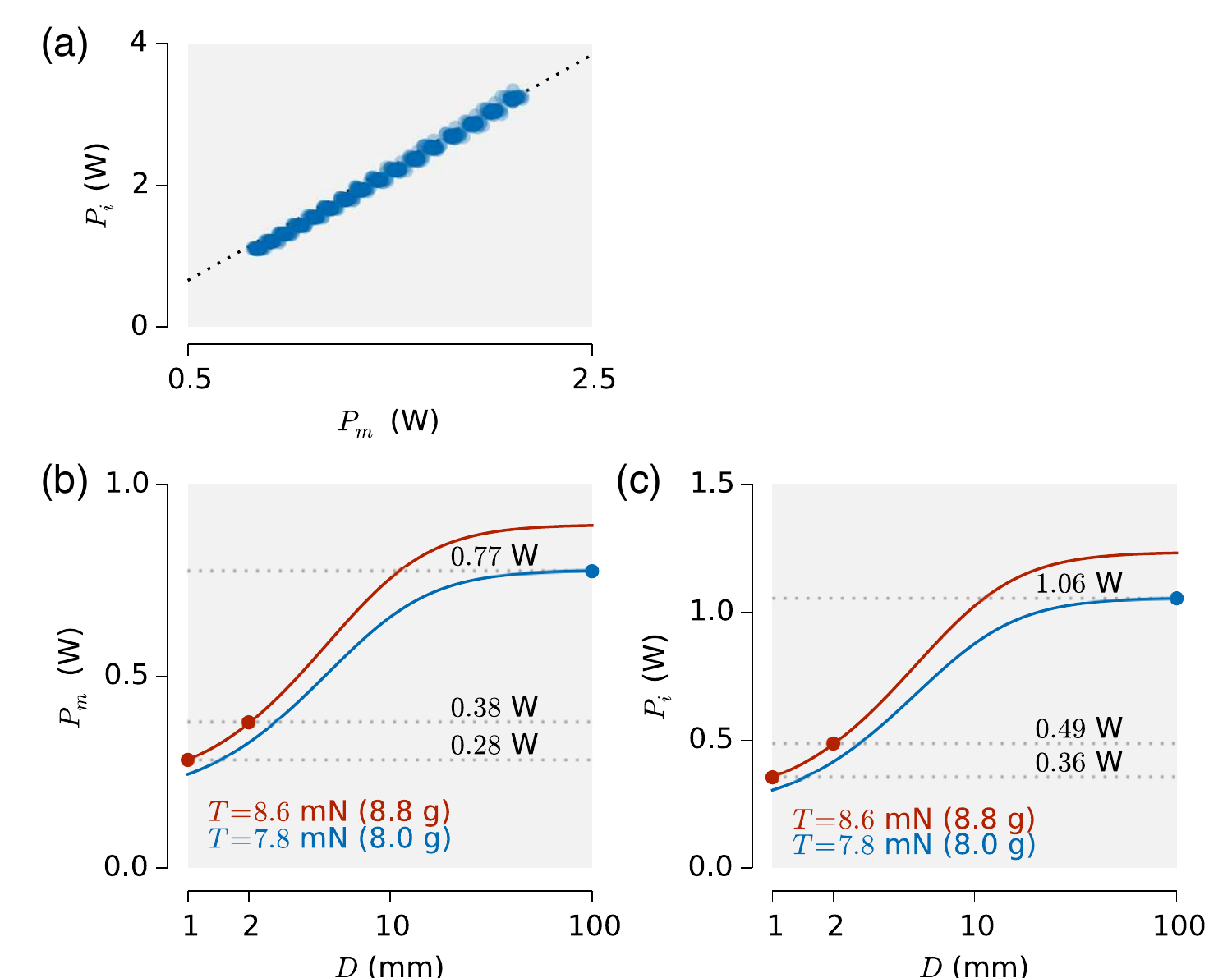}
\par\end{centering}
\caption{(a) Measured mechanical power and input power (blue points) and the
model prediction (dotted line). (b) The plot of the mechanical power
required to sustain $7.8$ mN (blue) and $8.6$ mN (red) thrust at
different ceiling distances. (c) The plot of the input power required
to sustain $7.8$ mN (blue) and $8.6$ mN (red) thrust at different
ceiling distances.}
\label{fig:panel_power_saving}
\end{figure}

\subsection{Power saving potential}

The analysis of the ceiling effects provided in this work relates
the thrust force to the aerodynamic power. Together with the assumption
that the aerodynamic power is proportional to the mechanical power
($P_{m}=\eta P_{a}$), it has been shown that the reduction in mechanical
power is given by the ceiling coefficient. In an actual robot, the
input power provided to the motor-propeller pair, however, is different
from the mechanical power as it includes frictional and dissipative
losses. In order to estimate the realistic power saving from the ceiling
effects, the analysis is expanded to include the input power for the
case of 23-mm propellers in a quadrotor configuration with the motor-to-motor
distance $L=92$ mm\textemdash identical to the configuration of a
commercially available Crazyflie 2.0 nanoquadrotor. The estimates
of input power, therefore, provide more realistic numbers for calculating
how much power saving could be achieved in practice.

The input power analysis begins by revisiting the first-order brushed
motor model: $V_{i}=IR_{i}+k\Omega$, the definitions of $P_{m}$
as $P_{m}=\tau\Omega=Ik\Omega$, and the torque coefficient: $c_{\tau}=\tau/\Omega^{2}$
. From earlier findings in figure \ref{fig:panel_model_flight_coeffs}(c),
it is reasonable to assume that $c_{\tau}$ is approximately constant,
regardless of the distance to the ceiling. In such circumstances,
we modify the motor model to represent the input and mechanical power
as
\begin{equation}
P_{i}=c_{\tau}^{2/3}k^{-2}R_{i}\cdot P_{m}^{4/3}+P_{m}.\label{eq:power_saving_potential}
\end{equation}
Using the same identified parameters from the experiments, $R=1.58$
$\Omega$, $k=1.1$ mV.s.rad$^{-1}$, and assuming a constant $c_{\tau}$
of $175$ nN.mm.s$^{2}$.rad$^{-2}$, the predicted relationship between
$P_{m}$ and $P_{i}$ for the system is shown in figure \ref{fig:panel_power_saving}(a).
The plot verifies that equation (\ref{eq:power_saving_potential})
matches the experimental data with reasonable accuracy.

According to the datasheet, the original Crazyflie 2.0 weighs 28 grams.
For practical flight with some safety margin, each propeller is required
to generate approximately an equivalent of $8$ grams of thrust force
(or $T=7.8$ mN) for the robot to hover. The required mechanical power
for the respective thrust is computed from equation (\ref{eq:mt_power_thrust}).
With the model of ceiling coefficients for this particular robot configuration
(figure \ref{fig:panel_ceiling_coefs}(e)), the mechanical power at
different ceiling-to-propeller distances needed to generate $T=7.8$
mN are shown in figure \ref{fig:panel_power_saving}(b). Furthermore,
the mechanical power is translated into the input power and shown
in figure \ref{fig:panel_power_saving}(c). The plots suggest that
the mechanical and input powers required for each propeller in a near
hovering condition for the robot are approximately $0.77$ W and $1.06$
W.

\begin{figure}
\begin{centering}
\includegraphics[width=8.5cm]{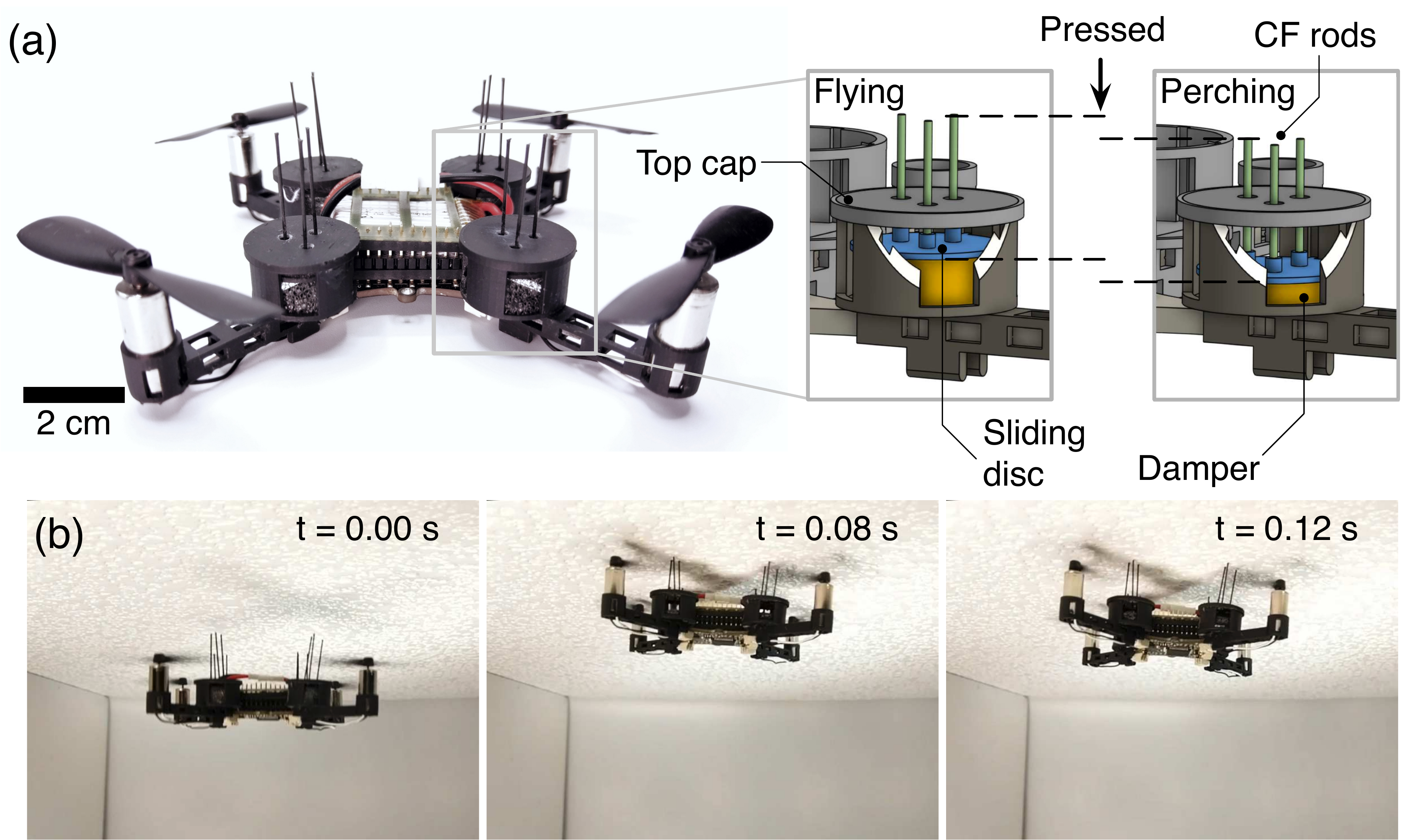}
\par\end{centering}
\caption{(a) A proof-of-concept prototype. The modified airframe features a
damping component that prevents the propellers from crashing directly
into a ceiling using carbon fiber rods. The kinetic energy is dissipated
with the incorporation of viscoelastic material. (b) Image sequence
showing the transition from aerial to surface locomotion.}
\label{fig:panel_perching}
\end{figure}
To leverage the ceiling effects for surface locomotion, we believe
it is necessary for the robot to generate the total thrust with a
greater magnitude than its weight such that there remains some normal
force against the ceiling. This normal force ensures contact between
the robot and the ceiling, and, therefore, maintains a constant separation.
On the assumption that the robot requires an additional of $10\%$
on thrust $(T=8.6$ mN or $8.8$ grams), the corresponding powers
for different ceiling distances are illustrated in figure \ref{fig:panel_power_saving}(b)-(c).
The calculation shows that the required mechanical and input powers
for each propeller are $0.38$ W and $0.49$ W when $D=2.0$ mm or
$0.28$ W and $0.36$ W when $D=1.0$ mm. From these estimates, the
use of ceilings for surface locomotion has the potential to reduce
power consumption of the robot by a factor of $2-3$. While these
numbers only account for the actuation and do not reflect the power
expended in the controller or communication, the high energetic requirement
of flight at small scales means the actuation power constitutes a
major proportion of the total power.

\subsection{Conceptual perching demonstration}

As stated, to demonstrate a ceiling locomotion of a small flying
robot, it necessitates a revision in mechanical designs of the robot
and a modification in the flight controller. Here, we present a preliminary
design of the mechanism that (i) keeps the propeller-to-ceiling distance
constant during perching; (ii) absorbs kinetic energy upon impact;
and (iii) is lightweight.

Using the Crazyflie 2.0 as a platform, the airframe is redesigned
with a damping element situated on each arm as shown in figure \ref{fig:panel_perching}(a).
The damping module is a hollow plastic cylinder, with a snap-on top
cap (in false-color green). The damping mechanism consists of a viscoelastic
element (false-color yellow, damper) sitting at the bottom, followed
by a sliding disc (false-color blue) with attached carbon fiber rods
(false-color purple, CF). The CF rods act as a structure preventing
the propellers from directly colliding with the ceiling. Upon an impact
with the ceiling, the CF rods and the base disc recede down along
the guided path and compress the viscoelastic component. The dissipated
energy reduces the bounces observed in flight tests. 

This initial design can be further tailored in future work. Currently,
with the added structure, the weight of the robot modestly increases
from 28 g to 30 g. As a proof-of-concept device, we show that this
simple mechanism is sufficient for a robot to perch on an overhang
in a human-operated flight test as presented in figure \ref{fig:panel_perching}(b). 

\section*{Acknowledgement}

This work was supported by the Research Grants Council of the Hong
Kong Special Administrative Region of China (grant numbers CityU-11205419
and CityU-11215117).

\bibliographystyle{ieeetr}
\bibliography{ms}

\begin{thebibliography}{10}

\bibitem{thurrowgood2014biologically}
S.~Thurrowgood, R.~J. Moore, D.~Soccol, M.~Knight, and M.~V. Srinivasan, ``A
  biologically inspired, vision-based guidance system for automatic landing of
  a fixed-wing aircraft,'' {\em Journal of Field Robotics}, vol.~31, no.~4,
  pp.~699--727, 2014.

\bibitem{oettershagen2016perpetual}
P.~Oettershagen, A.~Melzer, T.~Mantel, K.~Rudin, T.~Stastny, B.~Wawrzacz,
  T.~Hinzmann, A.~Kostas, and R.~Y. Siegwart, ``Perpetual flight with a small
  solar-powered uav: Flight results, performance analysis and model
  validation,'' in {\em Proceedings of the 2016 IEEE Aerospace Conference (AERO
  2016)}, p.~7500855, IEEE, 2016.

\bibitem{mulgaonkar2018robust}
Y.~Mulgaonkar, A.~Makineni, L.~Guerrero-Bonilla, and V.~Kumar, ``Robust aerial
  robot swarms without collision avoidance,'' {\em IEEE Robotics and Automation
  Letters}, vol.~3, no.~1, pp.~596--603, 2018.

\bibitem{vasarhelyi2018optimized}
G.~V{\'a}s{\'a}rhelyi, C.~Vir{\'a}gh, G.~Somorjai, T.~Nepusz, A.~E. Eiben, and
  T.~Vicsek, ``Optimized flocking of autonomous drones in confined
  environments,'' {\em Science Robotics}, vol.~3, no.~20, p.~eaat3536, 2018.

\bibitem{chirarattananon2016perching}
P.~Chirarattananon, K.~Y. Ma, and R.~J. Wood, ``Perching with a robotic insect
  using adaptive tracking control and iterative learning control,'' {\em The
  International Journal of Robotics Research}, vol.~35, no.~10, pp.~1185--1206,
  2016.

\bibitem{chen2017biologically}
Y.~Chen, H.~Wang, E.~F. Helbling, N.~T. Jafferis, R.~Zufferey, A.~Ong, K.~Ma,
  N.~Gravish, P.~Chirarattananon, M.~Kovac, {\em et~al.}, ``A biologically
  inspired, flapping-wing, hybrid aerial-aquatic microrobot,'' {\em Science
  Robotics}, vol.~2, no.~11, p.~eaao5619, 2017.

\bibitem{floreano2015science}
D.~Floreano and R.~J. Wood, ``Science, technology and the future of small
  autonomous drones,'' {\em Nature}, vol.~521, no.~7553, pp.~460--466, 2015.

\bibitem{karydis2017energetics}
K.~Karydis and V.~Kumar, ``Energetics in robotic flight at small scales,'' {\em
  Interface focus}, vol.~7, no.~1, p.~20160088, 2017.

\bibitem{duhamel2013biologically}
P.-E.~J. Duhamel, N.~O. P{\'e}rez-Arancibia, G.~L. Barrows, and R.~J. Wood,
  ``Biologically inspired optical-flow sensing for altitude control of
  flapping-wing microrobots,'' {\em IEEE/ASME Transactions on Mechatronics},
  vol.~18, no.~2, pp.~556--568, 2013.

\bibitem{chirarattananon2018direct}
P.~Chirarattananon, ``A direct optic flow-based strategy for inverse flight
  altitude estimation with monocular vision and imu measurements,'' {\em
  Bioinspiration \& biomimetics}, vol.~13, no.~3, p.~036004, 2018.

\bibitem{oktay2015comfortable}
T.~Oktay and C.~Sultan, ``Comfortable helicopter flight via passive/active
  morphing,'' {\em IEEE Transactions on Aerospace and Electronic Systems},
  vol.~51, no.~4, pp.~2876--2886, 2015.

\bibitem{morbidi2016minimum}
F.~Morbidi, R.~Cano, and D.~Lara, ``Minimum-energy path generation for a
  quadrotor uav,'' in {\em 2016 IEEE International Conference on Robotics and
  Automation (ICRA)}, pp.~1492--1498, IEEE, 2016.

\bibitem{gnemmi2017conception}
P.~Gnemmi, S.~Changey, K.~Meder, E.~Roussel, C.~Rey, C.~Steinbach, and
  C.~Berner, ``Conception and manufacturing of a projectile-drone hybrid
  system,'' {\em IEEE/ASME Transactions on Mechatronics}, vol.~22, no.~2,
  pp.~940--951, 2017.

\bibitem{daler2015bioinspired}
L.~Daler, S.~Mintchev, C.~Stefanini, and D.~Floreano, ``A bioinspired
  multi-modal flying and walking robot,'' {\em Bioinspiration \& biomimetics},
  vol.~10, no.~1, p.~016005, 2015.

\bibitem{alzu2018loon}
H.~Alzu'bi, I.~Mansour, and O.~Rawashdeh, ``Loon copter: Implementation of a
  hybrid unmanned aquatic--aerial quadcopter with active buoyancy control,''
  {\em Journal of Field Robotics}, 2018.

\bibitem{siddall2017fast}
R.~Siddall and M.~Kovac, ``Fast aquatic escape with a jet thruster,'' {\em
  IEEE/ASME Transactions on Mechatronics}, vol.~22, no.~1, pp.~217--226, 2017.

\bibitem{roderick2017touchdown}
W.~R. Roderick, M.~R. Cutkosky, and D.~Lentink, ``Touchdown to take-off: at the
  interface of flight and surface locomotion,'' {\em Interface focus}, vol.~7,
  no.~1, p.~20160094, 2017.

\bibitem{kalantari2015autonomous}
A.~Kalantari, K.~Mahajan, D.~Ruffatto, and M.~Spenko, ``Autonomous perching and
  take-off on vertical walls for a quadrotor micro air vehicle,'' in {\em
  Robotics and Automation (ICRA), 2015 IEEE International Conference on},
  pp.~4669--4674, IEEE, 2015.

\bibitem{graule2016perching}
M.~Graule, P.~Chirarattananon, S.~Fuller, N.~Jafferis, K.~Ma, M.~Spenko,
  R.~Kornbluh, and R.~Wood, ``Perching and takeoff of a robotic insect on
  overhangs using switchable electrostatic adhesion,'' {\em Science}, vol.~352,
  no.~6288, pp.~978--982, 2016.

\bibitem{pope2017multimodal}
M.~T. Pope, C.~W. Kimes, H.~Jiang, E.~W. Hawkes, M.~A. Estrada, C.~F. Kerst,
  W.~R. Roderick, A.~K. Han, D.~L. Christensen, and M.~R. Cutkosky, ``A
  multimodal robot for perching and climbing on vertical outdoor surfaces,''
  {\em IEEE Transactions on Robotics}, vol.~33, no.~1, pp.~38--48, 2017.

\bibitem{hang2019perching}
K.~Hang, X.~Lyu, H.~Song, J.~A. Stork, A.~M. Dollar, D.~Kragic, and F.~Zhang,
  ``Perching and resting-a paradigm for uav maneuvering with modularized
  landing gears,'' {\em Science Robotics}, vol.~4, no.~28, p.~eaau6637, 2019.

\bibitem{betz1937ground}
A.~Betz, ``The ground effect on lifting propellers,'' 1937.

\bibitem{griffiths2005predictions}
D.~A. Griffiths, S.~Ananthan, and J.~G. Leishman, ``Predictions of rotor
  performance in ground effect using a free-vortex wake model,'' {\em Journal
  of the American Helicopter Society}, vol.~50, no.~4, pp.~302--314, 2005.

\bibitem{leishman2006principles}
G.~J. Leishman, {\em Principles of helicopter aerodynamics with CD extra}.
\newblock Cambridge university press, 2006.

\bibitem{davis2016passive}
E.~Davis and P.~E. Pounds, ``Passive position control of a quadrotor with
  ground effect interaction.,'' {\em IEEE Robotics and Automation Letters},
  vol.~1, no.~1, pp.~539--545, 2016.

\bibitem{powers2013influence}
C.~Powers, D.~Mellinger, A.~Kushleyev, B.~Kothmann, and V.~Kumar, ``Influence
  of aerodynamics and proximity effects in quadrotor flight,'' in {\em
  Experimental robotics}, pp.~289--302, Springer, 2013.

\bibitem{hsiao2018ceiling}
Y.~H. Hsiao and P.~Chirarattananon, ``Ceiling effects for surface locomotion of
  small rotorcraft,'' in {\em Intelligent Robots and Systems (IROS), 2018
  IEEE/RSJ International Conference on}, IEEE, 2018.
\newblock to appear.

\bibitem{seddon2011basic}
J.~M. Seddon and S.~Newman, {\em Basic helicopter aerodynamics}, vol.~40.
\newblock John Wiley \& Sons, 2011.

\bibitem{bangura2016aerodynamics}
M.~Bangura, M.~Melega, R.~Naldi, and R.~Mahony, ``Aerodynamics of rotor blades
  for quadrotors,'' {\em arXiv preprint arXiv:1601.00733}, 2016.

\bibitem{bangura2017thrust}
M.~Bangura and R.~Mahony, ``Thrust control for multirotor aerial vehicles,''
  {\em IEEE Transactions on Robotics}, vol.~33, no.~2, pp.~390--405, 2017.

\bibitem{branlard2017wind}
E.~Branlard, {\em Wind Turbine Aerodynamics and Vorticity-Based Methods}.
\newblock Springer, 2017.

\bibitem{light1993tip}
J.~S. Light, ``Tip vortex geometry of a hovering helicopter rotor in ground
  effect,'' {\em Journal of the American helicopter society}, vol.~38, no.~2,
  pp.~34--42, 1993.

\bibitem{nathan2010rotor}
N.~D. Nathan, {\em The rotor wake in ground effect and its investigation in a
  wind tunnel}.
\newblock PhD thesis, University of Glasgow, 2010.

\end{thebibliography}

\cleardoublepage\newpage{}

\section*{Supplemental Materials}

\subsection*{Blade element method with radial flow}

\setcounter{page}{1}Traditional blade-element theory for a spinning
propeller assumes the flow through the propeller to be primarily along
the spinning direction and the propeller axis. According to our analysis
using momentum theory and the inviscid flow model, the presence of
a ceiling induces the flow along the radial direction. Here, we show
a brief derivation of equation (\ref{eq:bem_ceiling_thrust}) following
closely the approach used in \cite{bangura2016aerodynamics}.

\begin{figure}[h]
\begin{centering}
\includegraphics[width=4cm]{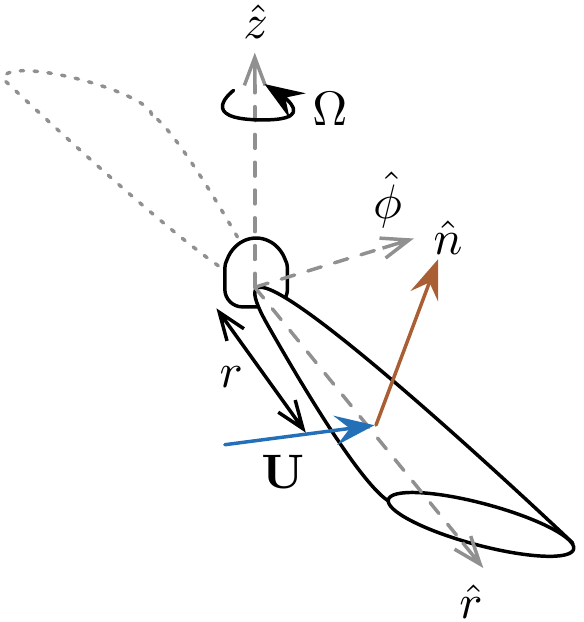}
\par\end{centering}
\caption{A spinning propeller with a propeller-attached frame. Vectors $\mathbf{U}$
indicates a relative local air velocity on the surface of the blade,
described by the normal vector $\hat{n}$, at distance $r$ from the
rotational axis.}
\label{fig:supp_bem}
\end{figure}
According to \cite{bangura2016aerodynamics}, the elemental thrust
of a spinning propeller at location $r$ from the spinning axis is
\begin{equation}
\mathrm{d}T\left(r\right)\approx\frac{1}{2}\rho U\left(r\right)^{2}C_{L}\left(\alpha\right)c\left(r\right)\mathrm{d}r,\label{eq:supp_thrust_element}
\end{equation}
where $U\left(r\right)$ is the relative air speed with respect to
the propeller, $C_{L}\left(\alpha\right)$ is the lift coefficient
with the angle of attack $\alpha$, and $c\left(r\right)$ is the
propeller chord. The magnitude of $U\left(r\right)$ is calculated
from a vector sum of three perpendicular velocity components: the
velocity along the spinning direction $U_{\Omega}=\Omega r$, the
vertical direction $U_{z}$, and the radial direction $U_{r}$, where
it is assumed that $U_{\Omega}\gg U_{z},U_{r}$. Let vectors $\hat{z}$,
$\hat{r}$, and $\hat{\phi}$ describe propeller-attached cylindrical
coordinates as shown in figure \ref{fig:supp_bem}. If $\mathbf{U}$
is a vector representative of $U\left(r\right)$, such that $\mathbf{U}=\left[\begin{array}{ccc}
U_{r} & U_{z} & U_{\Omega}\end{array}\right]^{T}$, then $\alpha$ is the angle between $\mathbf{U}\left(r\right)$
and the unit vector normal to the local propeller surface, $\mathrm{\hat{n}}\left(r\right)$,
as illustrated in figure \ref{fig:supp_bem}. For a flat fixed-pitch
propeller, we assume the local profile can be described by two small
angles such that $\hat{n}\left(r\right)=\left[\begin{array}{ccc}
\sin\theta_{\phi} & \cos\theta_{r}\cos\theta_{\phi} & \sin\theta_{r}\cos\theta_{\phi}\end{array}\right]^{T}$. The angle of attack, using the small angle approximation, is given
by $\sin\alpha=\mathbf{U}\cdot\hat{n}/\left\Vert \mathbf{U}\right\Vert \approx\sin\left(\theta_{r}+\arcsin\left(U_{z}/U\right)\right)+\frac{U_{r}}{U}\sin\theta_{\phi}$,
or 
\begin{equation}
\alpha\left(r\right)\approx\theta_{r}\left(r\right)+\frac{U_{z}\left(r\right)}{U}+\theta_{\phi}\left(r\right)\frac{U_{r}\left(r\right)}{U}.\label{eq:supp_aoa}
\end{equation}
The lift coefficient is further approximated based on the small angle
of attack assumption as $C_{L}\left(\alpha\right)\approx C_{L\alpha}\alpha$.
As a consequence, equation (\ref{eq:supp_thrust_element}) becomes
\begin{align}
\mathrm{d}T\left(r\right) & \approx\frac{1}{2}\rho C_{L\alpha}U_{\Omega}^{2}\left(\theta_{r}\left(r\right)+\frac{U_{z}\left(r\right)}{U}\right.\label{eq:supp_thrust_element_final}\\
 & \left.+\theta_{\phi}\left(r\right)\frac{U_{r}\left(r\right)}{U}\right)c\left(r\right)\mathrm{d}r.\nonumber 
\end{align}
Referring to the previous notations from section \ref{sec:MomentumTheory},
we can substitute $U$ and $U_{\Omega}$ with $r\Omega$, $U_{z}$
with the induced velocity (-$v_{i}$) and $U_{r}\left(r\right)$ with
$v_{r}\left(r\right)=\frac{r}{2D}v_{i}$. Depending on the geometrical
profile of the propeller, the integration of equation (\ref{eq:supp_thrust_element_final})
can be expressed as
\begin{align}
\mathrm{d}T\left(r\right) & \approx\frac{1}{2}\rho C_{L\alpha}\left(\Omega^{2}r^{2}\theta_{r}\left(r\right)-v_{i}\Omega r\right.\nonumber \\
 & \left.+\theta_{\phi}\left(r\right)\frac{rv_{i}}{2D}\Omega r\right)c\left(r\right)\mathrm{d}r.\label{eq:supp_thrust_element_f2}
\end{align}
using three lumped parameters ($c_{0}$, $c_{1}$, $c_{2}$):
\begin{equation}
T=\frac{1}{2}\rho AR^{2}(c_{0}-c_{1}\frac{v_{i}}{\Omega R}+c_{2}\frac{v_{i}}{\Omega R}\delta)\Omega^{2},\label{eq:supp_balde_element_final}
\end{equation}
where
\begin{align}
c_{0} & =\frac{C_{L\alpha}}{AR^{2}}\int_{r=0}^{R}c\left(r\right)\theta_{r}\left(r\right)r^{2}\mathrm{d}r\nonumber \\
c_{1} & =\frac{C_{L\alpha}}{AR^{2}}\int_{r=0}^{R}Rc\left(r\right)r\mathrm{d}r\nonumber \\
c_{2} & =\frac{C_{L\alpha}}{AR^{2}}\int_{r=0}^{R}\frac{c\left(r\right)}{2}\theta_{\phi}\left(r\right)r^{2}dr.\label{eq:supp_thrust_element_f3}
\end{align}
Without radial flow, equation (\ref{eq:supp_thrust_element_f3}) reduces
to
\[
T=\frac{1}{2}\rho AR^{2}(c_{0}-c_{1}\frac{v_{i}}{\Omega R})\Omega^{2},
\]
similar to the result from \cite{bangura2016aerodynamics} after some
algebraic manipulation.

\subsection*{Mechanical resonance}

In figure \ref{fig:panel_ceiling_coefs}, we observe irregular drops
in the values of the ceiling coefficients uncaptured by the proposed
model. These dips are more evident in the case of the 50-mm propeller
(at $\delta=7.1$) and some configurations of 23-mm propellers (at
$\delta=9.2$). During the experiments, we noticed unusually loud
noises produced by the system around these points. We believe this
could be caused by a condition that promotes resonance that results
in some vibration and undesired power loss. This is consistent with
the reduction in the values of the ceiling coefficient, which indicates
the unmodeled reduced power efficiency of the system. In an attempt
to explain the phenomenon, here,we offer some physics-based explanations. 

In our system, the only excitation frequency stems from the spinning
propeller, and the resonance was observed across multiple frequencies.
The fact that the drops in $\gamma$ occur at some specific $\delta$
(or at a fixed $D$), which is deduced from the relationship between
$T$ and $P_{m}$ as given in equation (\ref{eq:mt_power_thrust}),
means that the the reduced power efficiency happens when the propellers
spin at various rotational rates. This suggests that the events are
(i) not related to a resonant frequency of any particular mechanical
structures of the system; and (ii) not a result of standing sound
waves as there would be a single resonant frequency for a particular
value of $D$.

On the other hand, we propose that the resonance is related to stationary
waves caused by the airflow, such that the wave speed is related to
the flow speed, or induced velocity, $(v\sim v_{i}$). If the wavelength
is proportional to the separation $D$, then the stationary waves
must satisfy the condition $v=\lambda f$, or
\begin{equation}
v_{i}\sim Df.\label{eq:supp_res_wave}
\end{equation}
We re-write the solution of equations (\ref{eq:bem_ceiling_thrust})
and (\ref{eq:bem_thrust_vi}) as 
\begin{align*}
v_{i}/\Omega R & =\frac{1}{2c_{0}}\left[\left(c_{1}-c_{2}\delta\right)+\sqrt{\left(c_{1}-c_{2}\delta\right)^{2}+16c_{0}\gamma^{2}}\right]\\
 & =g\left(\delta\right).
\end{align*}
When substituted into the equation above, this yields
\begin{align}
g\left(\delta\right)\Omega R & \sim Df\nonumber \\
2\pi\delta g\left(\delta\right) & \sim\mbox{constant},\label{eq:supp_res_scaling}
\end{align}
where we have used $\Omega=2\pi f$. The result suggests that, there
should be a particular value of $\delta g\left(\delta\right)$ that
satisfies the stationary wave conditions. Inspection of figure \ref{fig:panel_ceiling_coefs}
reveals that there are three configurations corresponding to figure
\ref{fig:panel_ceiling_coefs}(b), (e), and (f) with noticeable reduction
of $\gamma$'s at $\delta=7.2$, $9.2$, and $9.2$. From our fitted
model parameters, the values of $\delta g\left(\delta\right)$ for
these configurations are: 0.6317 , 0.6726, 0.7295. That is, they are
within $15\%$ from one another regardless of the propeller size,
supporting our analysis which suggests that they should be a single
constant in theory, independent of the propeller radius.
\end{document}